\documentclass[10pt,twocolumn,letterpaper]{article}
\usepackage{subfig}
\usepackage{float}
\usepackage{iccv}
\usepackage{times}
\usepackage{epsfig}
\usepackage{amsmath}
\usepackage{amssymb}
\usepackage{threeparttable}
\usepackage[mathscr]{euscript}

\usepackage{amsmath, bm}
\usepackage{multirow}
\usepackage{graphicx}
\usepackage{booktabs}
\usepackage{gensymb}
\usepackage{hyperref}

\usepackage{verbatim}

\renewcommand{\texttt}[1]{ $ {{\tt #1} } $}

\usepackage[margin=4pt,font=footnotesize,labelfont=bf,labelsep=endash,tableposition=top]{caption}

\iccvfinalcopy %

\ificcvfinal\fi
\begin{document}

\title{Enforcing geometric constraints of virtual normal for depth prediction}

\author{
 Wei Yin$ ^1$ ~~~~
 Yifan Liu$ ^1$ ~~~~
 Chunhua Shen$ ^1$\thanks{Corresponding author, email: \texttt{chunhua.shen@adelaide.edu.au}}
  ~~~~
 Youliang Yan$ ^2$
 \\ $ ^1 $The University of Adelaide, Australia
 ~~~ ~ ~
 $ ^2 $Noah's Ark Lab, Huawei Technologies
}

\maketitle
\begin{abstract}
Monocular depth prediction plays a crucial role in understanding $3$D scene geometry. Although recent methods have achieved impressive progress in
evaluation metrics such as the pixel-wise relative error, most methods neglect the geometric constraints in the 3D space. In this work, we show the importance of the high-order 3D geometric constraints for  depth prediction. By designing a loss term that enforces one simple type of geometric constraints, namely, \emph{virtual normal} directions determined by randomly sampled three points in the reconstructed 3D space,
we can considerably improve the depth prediction accuracy.
Significantly, the byproduct of this predicted depth being sufficiently accurate is that we are now able to recover  good  3D structures of the scene such as the point cloud and surface normal
directly from the depth,
eliminating the necessity of training new sub-models as was previously done.
Experiments on two %
benchmarks: NYU Depth-V2
and KITTI
demonstrate
the effectiveness of our method and state-of-the-art performance.
Code is available at:

\url{https://tinyurl.com/virtualnormal}

\end{abstract}

\section{Introduction}
Monocular depth prediction aims to predict distances between scene objects and the camera  from  a single monocular image.
It is a critical task for understanding the 3D scene,  such as recognizing a 3D object and parsing a 3D scene.
\begin{figure}[!t]
\centering
\includegraphics[width=0.47\textwidth]{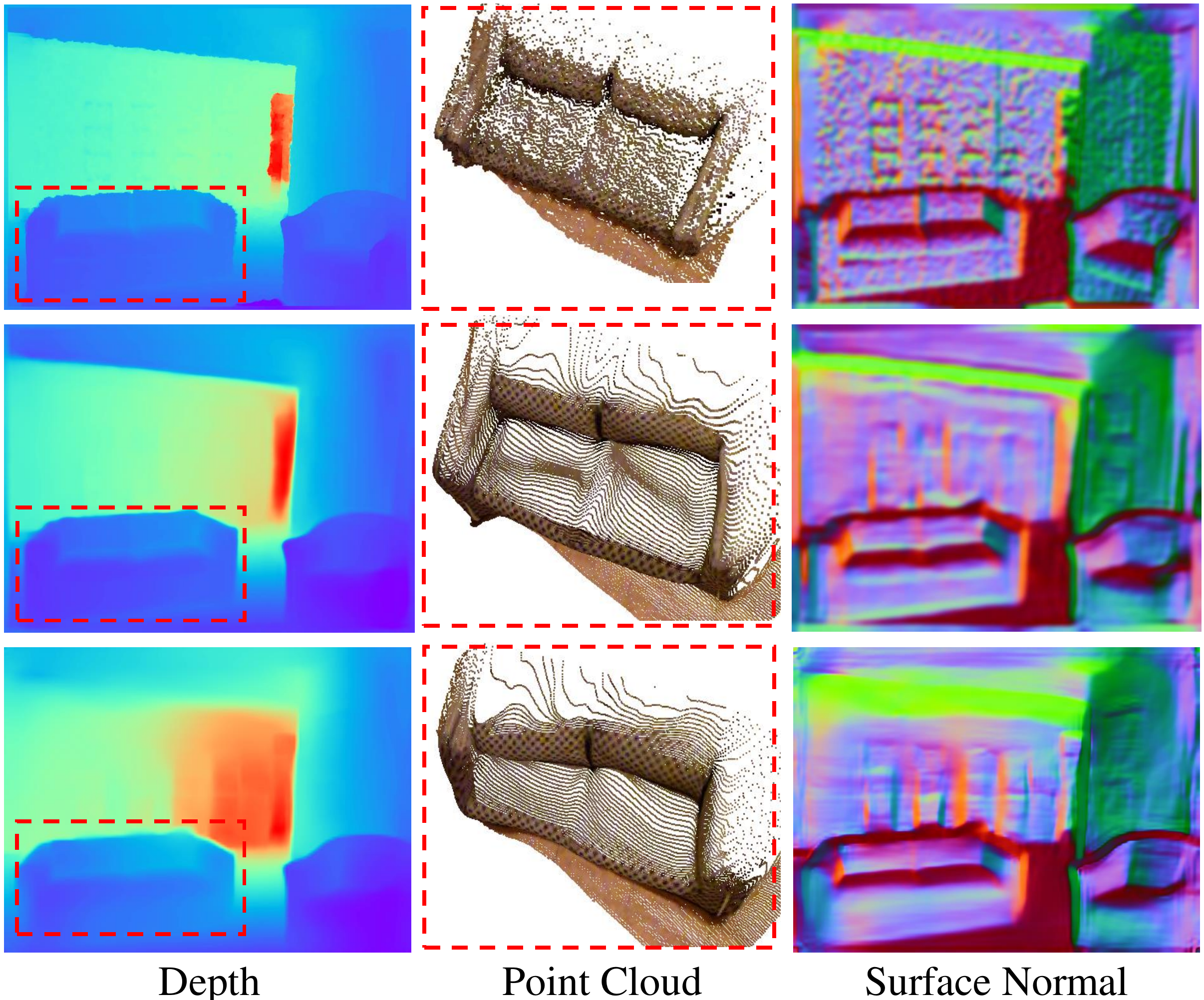}
\caption{Example results of ground truth (the first row), our method (the second row) and  Hu \etal \cite{Hu2018RevisitingSI} (the third row). By enforcing the geometric constraints of
virtual normals, %
our reconstructed 3D point cloud can represent better shape of sofa (see the left part) and the recovered surface normal has much less errors (see green parts) even though the absolute relative error (rel) of our predicted depth
is only slightly better than Hu \etal
($0.108$ \emph{vs.}\ $0.115$).
}
\label{fig:pcd normal dorn ours}
\vspace{-1em}
\end{figure}

Although the monocular depth prediction is
an ill-posed problem because many 3D scenes can be projected to the same 2D image, many deep convolutional neural networks (DCNN) based methods~\cite{eigen2015predicting,eigen2014depth, fu2018deep,guo2018learning,  laina2016deeper,   li2018deep, saxena2009make3d} have achieved impressive results by using a large amount of labelled data, thus taking advantage of prior knowledge in labelled data to solve the ambiguity.

These methods typically formulate the optimization problem as either point-wise regression or classification. That is, with the i.i.d.\ assumption, the overall loss is  summing over all pixels.
To improve the performance, some endeavours have been made to employ other constraints besides the pixel-wise term. For example, a continuous conditional random field (CRF)~\cite{liu2016learning} is used for depth prediction, which takes
pair-wise information into account. Other high-order
geometric relations~\cite{fei2018geo,qi2018geonet} are also exploited, such as designing a gravity constraint for local regions~\cite{fei2018geo} or incorporating the depth-to-surface-normal mutual transformation inside the optimization pipeline~\cite{qi2018geonet}.
Note that, for the above methods, almost all the geometric constraints are `local' in the sense that they are extracted from a small neighborhood in either 2D or 3D.
{Surface normal is `local' by nature as it is defined by  the local tangent plane. }
As the ground truth depth maps of most datasets are captured by consumer-level sensors, such as the Kinect, depth values can fluctuate considerably. Such noisy measurement would
adversely affect the precision and subsequently the effectiveness of those local constraints inevitably.
Moreover, local constraints calculated over a small neighborhood have not fully exploited the  structure information of the scene geometry that may be possibly used to boost the performance.

To address these limitations,  here we propose a more stable geometric constraint from a global perspective to
take
 long-range relations into account for predicting depth, termed  \emph{virtual normal}.
 A few previous methods already made use of
 3D geometric information in depth estimation, almost all of which
 focus  on using surface normal.
 \textit{We instead
 reconstruct the 3D point cloud from the estimated  depth map explicitly.  }
In other words, we generate the 3D scene
by lifting each RGB pixel in the 2D image to its corresponding 3D coordinate with the estimated depth map.
This 3D point cloud serves as an intermediate representation.
With the reconstructed point cloud, we can exploit many kinds of 3D geometry information,
 not limited to the surface normal.
Here we consider the long-range dependency in the 3D space by randomly sampling three non-colinear points with the large distance to form a \emph{virtual plane}, of which the normal vector is the proposed \emph{virtual normal} (VN). The direction
divergence
between ground-truth and predicted VN %
can serve as a high-order 3D geometry loss.
Owing to the long-range sampling of points,
the adverse impact caused by noises in depth measurement is much alleviated
compared to the computation of the surface normal, making VN  significantly more accurate.
Moreover,
with randomly  sampling we can obtain  a large number of such constraints,
encoding the global 3D geometric.
Second,\textit{ by converting estimated depth maps from images  to 3D point cloud representations it opens
many possibilities of incorporating  algorithms for 3D point cloud processing to 2D images and 2.5D depth processing.} Here we  show one instance of such possibilies.

By combining the high-order geometric supervision  and the pixel-wise
    depth supervision,  our  network can  predict not only an accurate depth map but also  the high-quality 3D point cloud, subsequently other geometry information such as
the surface normal.
It is worth noting that we do not use a new model or introduce network branches for
estimating the surface normal. Instead it is computed directly from the reconstructed point cloud.
The second row of Fig.~\ref{fig:pcd normal dorn ours} demonstrates an example of our results. By contrast, although the previously state-of-the-art method~\cite{Hu2018RevisitingSI} predicts the depth with low errors, the reconstructed point cloud
is far away from the
original shape (see, e.g., left part of `sofa'). The surface normal also contains many errors.
We are probably the first to achieve high-quality monocular depth and surface normal prediction with a single network.

Experimental results on NYUD-v2~\cite{silberman2012indoor} and KITTI~\cite{geiger2013vision} datasets demonstrate
state-of-the-art performance of our method.
Besides, when training with the lightweight backbone, MobileNetV2~\cite{sandler2018mobilenetv2}, our framework  provides a better trade-off between network parameters and accuracy. Our method outperforms other state-of-the-art real-time systems by up to $29$\% with a comparable number of network parameters. Furthermore, from the reconstructed point cloud,
we %
directly calculate the surface normal, with a precision being on par with that of
specific DCNN based surface normal estimation methods.
In summary,
our main contributions of this work are as follow.
\begin{itemize}
\itemsep -0.15cm
\item We demonstrate the effectiveness of enforcing a high-order geometric constraint in the 3D space for the depth prediction task.
Such  global geometry information is instantiated with a simple yet effective
concept termed \emph{virtual normal} (VN).
By enforcing a loss defined on VNs, we %
demonstrate
the importance of 3D geometry information in depth estimation,
and
design a simple loss to exploit it.

\item Our method can reconstruct high-quality 3D scene point clouds, from which other 3D geometry features can be  calculated, such as the surface normal.
In essence, we show that for depth estimation, one should not consider the information represented by depth only. Instead, converting depth into 3D point clouds and exploiting 3D geometry is likely to improve many tasks including depth estimation.
\item Experimental results on NYUD-V2 and KITTI illustrate that our method  achieves state-of-the-art performance.
\end{itemize}

\subsection{Related Work}
\noindent\textbf{Monocular Depth Prediction.} Depth prediction from images is a long-standing problem. Previous work can be divided into active methods and passive methods. The former ones use the assistant optical information for prediction, such as coded patterns~\cite{yin2017high}, while the latter ones completely focus on image contents.
Monocular depth prediction~\cite{cao2017estimating,eigen2015predicting,eigen2014depth,  liu2016learning,  zheng2018net} has been extensively studied recently. %
As limited geometric information can be directly extracted from the monocular image,
it is essentially an ill-posed problem.
Recently, owing to the structural features from very deep convolution neural network, such as ResNet~\cite{he2016deep}, various DCNN-based methods learn to predict depth with deep CNN features.
Fu \etal~\cite{fu2018deep} proposed an encoder-decoder network, which extracts multi-scale features from the encoder and is trained in an end-to-end manner without iterative refinement. They achieved state-of-the-art performance on several datasets.
Jiao~\etal\cite{jiao2018look} proposed an attention-driven loss, which merges the semantic priors to improve the prediction precision on unbalanced distribution datasets.

Most previous methods only adopted the pixel-wise depth
 supervision to train a network. By contrast, Liu~\etal \cite{liu2016learning} combined DCNN with the continuous conditional random field (CRF) to exploit consistency information of neighbouring pixels. CRF establishes a pair-wise constraint for local regions. Furthermore, several high-order constraints are investigated. Chen~\etal \cite{chen2018rethinking} applied the generative adversarial training to lead the network to learn a context-aware and patch-level loss automatically.
Note that most of these methods directly work with the depth, instead of in the 3D space.

\begin{figure*}[!ht]
\vspace{-1em}
\centering
\includegraphics[width=.82\textwidth]{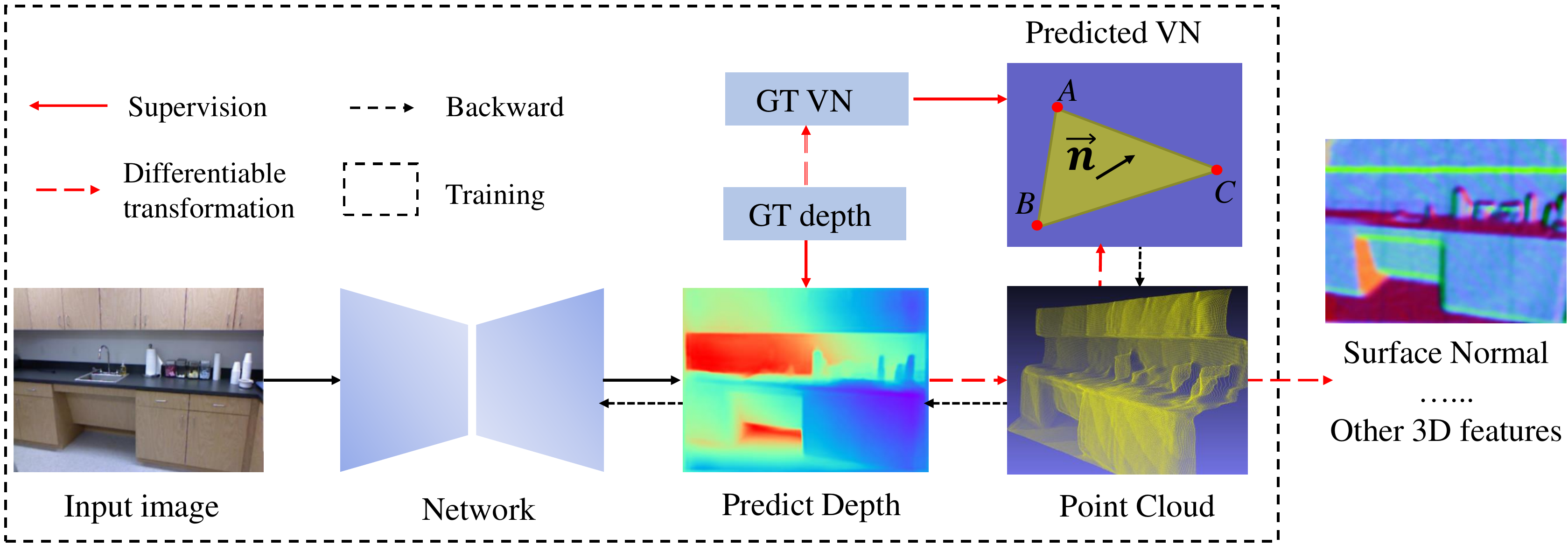}
\caption{Illustration of the pipeline of our method. An encoder-decoder network is employed to predict the depth, from which the point cloud can be reconstructed. A pixel-wise depth supervision is firstly enforced on the predicted depth, while a geometric supervision, virtual normal constraint, is enforced in 3D space.
With the well trained model, other 3D features, such as the surface normal, can be directly recovered from the reconstructed 3D point cloud in the inference.
}
\vspace{-1.2em}
\label{fig:architecture}
\end{figure*}

\noindent\textbf{Surface Normal.} Surface normal is an important geometry information for 3D scene understanding. Several data-driven methods~\cite{eigen2015predicting,eigen2014depth, fouhey2013data, fouhey2014unfolding,wang2015designing,   zeisl2014discriminatively} have achieved promising results. Eigen \etal~\cite{eigen2015predicting} proposed a CNN with different output channels to directly predict depth map, surface normal and semantic labels. Bansal \etal~\cite{bansal2016marr} proposed a two-stream network to predict the surface normal first, which is further joined with the input image to learn the pose. %
Note that most of these methods formulate surface normal prediction and depth prediction as multiple different tasks.

\section{Our Method}

Our approach resolves the monocular depth prediction and reconstructs the high-quality scene 3D point cloud from the predicted depth at the same time. The pipeline is illustrated in Fig.~\ref{fig:architecture}.

We take an RGB image $I_{in}$ as the input of an encoder-decoder network
and predict the depth map $D_{pred}$. From the $D_{pred}$, the 3D scene point cloud $P_{pred}$ can be %
reconstructed.
The ground truth point cloud $P_{gt}$ is reconstructed from  $D_{gt}$.

We enforce two
types
of supervision for training the network.%
 We firstly follow
 standard monocular depth prediction methods to enforce  pixel-wise depth supervision over $D_{pred}$ with
 $D_{gt}$.
 With the reconstructed point clouds, we then  align the spatial relationship between the $P_{pred}$ and the $P_{gt}$ using  the proposed \emph{virtual normal}.

When the network is well trained, we not only obtain accurate depth map but also high-quality point clouds. From the reconstructed point clouds, other 3D features can be
directly
calculated, such as the surface normal.

\subsection{High-order Geometric Constraints}

\begin{figure}[!bth]
\centering
\includegraphics[width=0.45\textwidth]{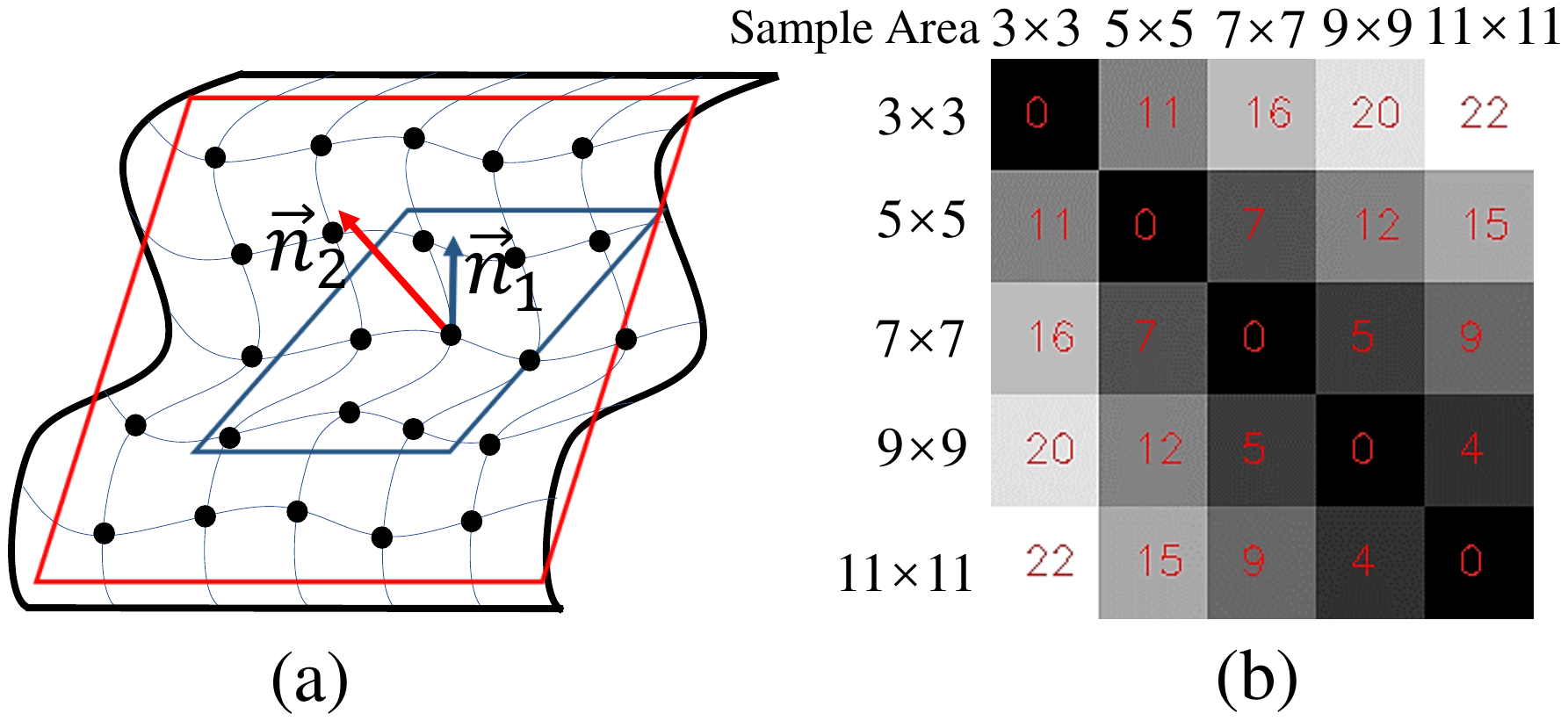}
\caption{Illustration of fitting point clouds to obtain the local surface normal. The directions of the surface normals is fitted with
different sampling sizes
on a real point cloud (a).
Because of noise, the surface normals vary significantly.
(b) compares the angular difference between surface normals computed with different sample sizes in %
Mean Difference Error.
The error  can vary significantly. }

\label{fig:fit plane}
\vspace{-2em}
\end{figure}

\noindent\textbf{Surface Normal.} The surface normal is an important `local' feature for many point-cloud based applications such as registration~\cite{rusu2008aligning} and object detection~\cite{hinterstoisser2011multimodal, gupta2014learning}.
It
appears to be
a promising 3D cue for improving  depth prediction.
One can apply the angular difference between ground-truth and calculated surface normal to be a geometric constraint. One major issue of this approach is,
when %
computing
 surface normal from either a depth map or  3D point cloud,
it is sensitive to noise. Moreover, surface normal only considers short-range local information.

We follow~\cite{klasing2009comparison} to calculate the surface normal. It assumes that local 3D points locate in the same plane, of which the normal vector is the surface normal. %
In practice ground-truth depth maps are usually captured by a consumer-level sensor with
limited  precision, so depth maps are contaminated by noise.
The reconstructed point clouds in the local region can vary considerably due to noises as well as the size of local patch for sampling (Fig.~\ref{fig:fit plane}(a)).
We experiment on the NYUD-V2 dataset to test the robustness of
 the surface normal computation.
Five different sampling sizes around the target pixel are employed to sample points, which are used to calculate its surface normal. The sample area is $a = (2i +1) \cdot (2i +1), i = 1, ..., 5$. The Mean Difference Error (Mean)~\cite{eigen2015predicting} between calculated surface normals is evaluated.
 From Fig.~\ref{fig:fit plane}(b), we can learn
 that the surface normal varies significantly with different sampling sizes. For example, the Mean between 3$ \times$3 and 11$\times$11 is 22\degree. Such unstable surface normal
negatively affects its effectiveness for learning.
Likewise,
other 3D geometric constraints demonstrating the `local' relative relations also %
encounter this problem.

\noindent\textbf{Virtual Normal.}
In order to enforce  robust high-order geometric supervision in the  3D space, we propose the virtual normal (VN) to establish 3D geometric connections between regions in a much larger range.
The point cloud can be reconstructed from the depth based on the pinhole camera model. For each pixel $p_{i}(u_{i}, v_{i})$, the 3D location $P_{i}(x_{i}, y_{i}, z_{i})$ in the world coordinate can be obtained by the prospective projection. We set the camera coordinate as the world coordinate. Then the 3D coordinate $P_{i}$ is denoted as follows:
\begin{equation}
\begin{split}
z_{i} = d_{i}, x_{i} = \frac{d_{i} \cdot (u_{i} - u_{0})}{f_{x}},y_{i} = \frac{d_{i}(v_{i} - v_{0})}{f_{y}}
\end{split}
\end{equation}
where $d_{i}$ is the depth. $f_{x}$ and $f_{y}$ are the focal length along the $x$ and $y$ coordinate axis respectively. $u_{0}$ and $v_{0}$ are the 2D coordinate of the optical center.

We randomly sample $N$ groups points from the depth map, with three points in each group. The corresponding 3D points are $\mathscr{S} = \{(P_{A}, P_{B}, P_{C})_{i} | i = 0...N\}$. Three points in a group are restricted to be non-colinear based on the restriction $\mathscr{R}_{1}$. $\angle(\cdot)$ is the angle between two vectors.
\begin{equation}
\begin{split}
    \mathscr{R}_{1} = \{\alpha \geq \angle(\overrightarrow{P_{A}P_{B}} , \overrightarrow{P_{A}P_{C}} )  \geq \beta, \\
    \alpha  \geq \angle(\overrightarrow{P_{B}P_{C}} , \overrightarrow{P_{B}P_{A}} )  \geq \beta | P\in \mathscr{S} \}
\end{split}
\end{equation}
where $\alpha, \beta$ are hyper-parameters. In all experiments, we set $ \alpha= 120\degree $, $ \beta = 30\degree $

In order to sample more long-range points, which have ambiguous relative locations in 3D space, we perform long-range restriction $\mathscr{R}_{2}$ for each group in $\mathscr{S}$.
\begin{equation}
    \mathscr{R}_{2} = \{\|\overrightarrow{P_{k}P_{m}}\|>\theta | k, m \in [A, B, C], P\in \mathscr{S} \}
\end{equation}
where $\theta =0.6m$ in our experiments.

Therefore, three 3D points in each group can establish a plane. We
compute the normal vector of the plane to
encode
geometric relations, which can be written as
\begin{equation}
\label{eq:normal}
\begin{split}
    \mathscr{N} = \{\boldsymbol{n_{i}}
    = \frac{\overrightarrow{P_{Ai}P_{Bi}}\times\overrightarrow{P_{Ai}P_{Ci}}}
    {\left \| \overrightarrow{P_{Ai}P_{Bi}}\times \overrightarrow{P_{Ai}P_{Ci}} \right \|} | \\
     {(P_{A}, P_{B}, P_{C})_{i} \in \mathscr{S}},
    {i = 0...N}\}
\end{split}
\end{equation}
where $\boldsymbol{n_{i}}$ is the normal vector of the virtual plane $i$.

\begin{figure}[!ht]
\centering
\includegraphics[width=0.25\textwidth]{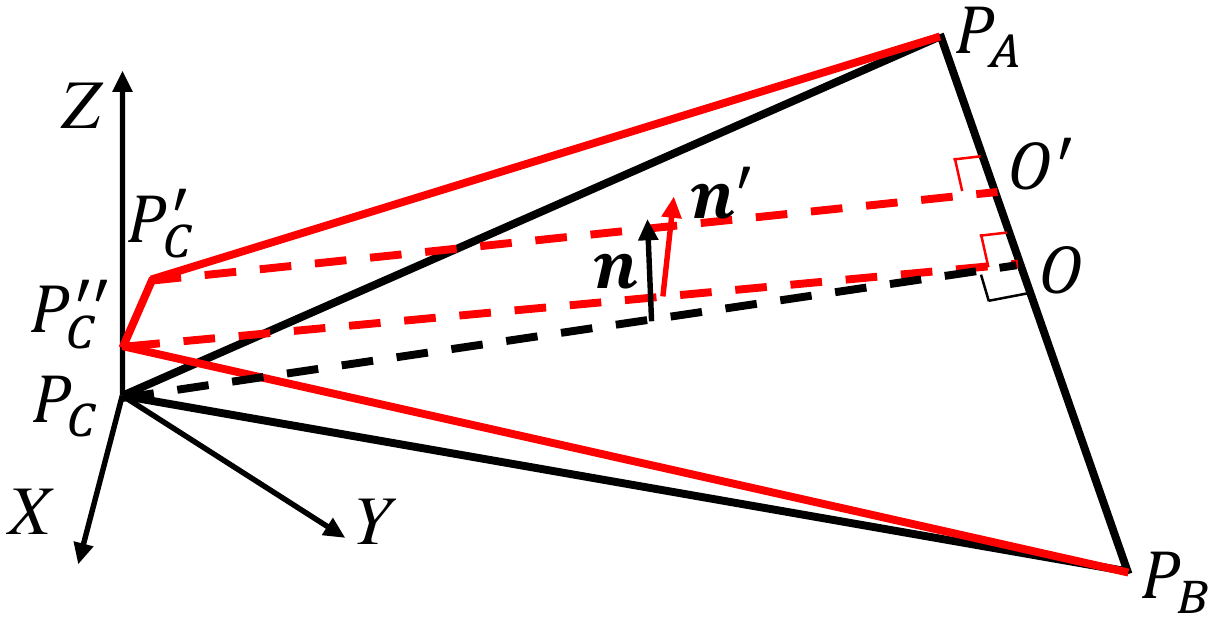}
\caption{Robustness of VN to depth noise.}
\label{fig:VNL noise}
\end{figure}

\noindent\textbf{Robustness to Depth Noise.}
Compared with local surface normal, our virtual normal is more robust to noise.
In Fig.~\ref{fig:VNL noise}, we sample three 3D points with large distance. $P_{A}$ and $P_{B}$ are assumed to locate on the $XY$ plane, $P_{C}$ is on the $Z$ axis. When $P_{C}$ varies to ${P_{C}}'$, the direction of the virtual normal changes from $\boldsymbol{n}$ to ${\boldsymbol{n}}'$. ${P_{C}}''$ is the intersection point between plane $P_{A}P_{B}{P_{C}}'$ and $Z$ axis. Because of restrictions $\mathscr{R}_{1}$ and $\mathscr{R}_{2}$, the difference between $\boldsymbol{n}$ and ${\boldsymbol{n}}'$ is
usually very small, which is simple to show:
\begin{equation}
\begin{split}
    \angle(\boldsymbol{n} ,{\boldsymbol{n}}') =
    &  \angle (\overrightarrow{OP_{C}}, \overrightarrow{O{P_{C}}''} ) = \arctan{\frac{\|\overrightarrow{P_{C}{P_{C}}''}\|}{\|\overrightarrow{OP_{C}}\|}} \approx 0,\\
    &\|\overrightarrow{P_{C}{P_{C}}''}\|
    \ll
    \|\overrightarrow{OP_{C}}\|
\end{split}
\end{equation}

\begin{figure}[bth!]
\centering    %
\subfloat[] %
{
	\includegraphics[width=0.13\textwidth]{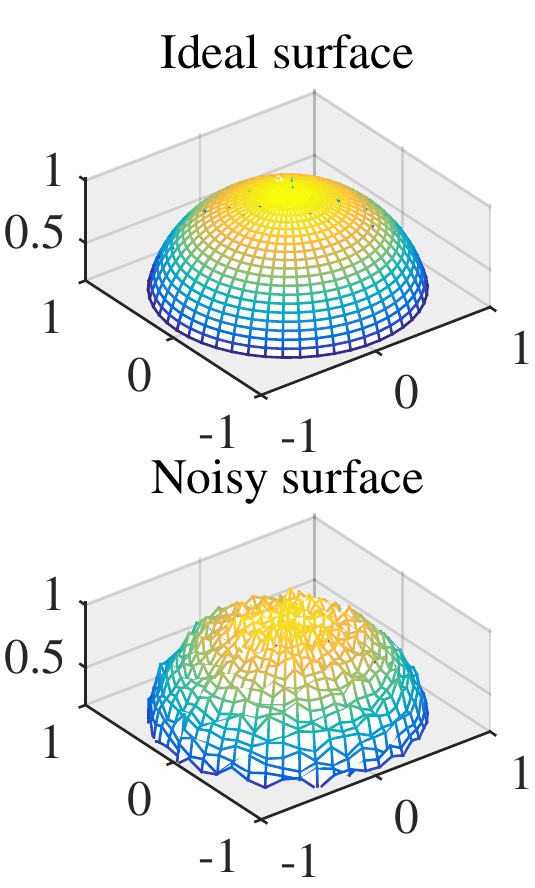}
	\label{fig:sn vn robustness sphere}
}
\subfloat[] %
{
	\includegraphics[width=0.32\textwidth]{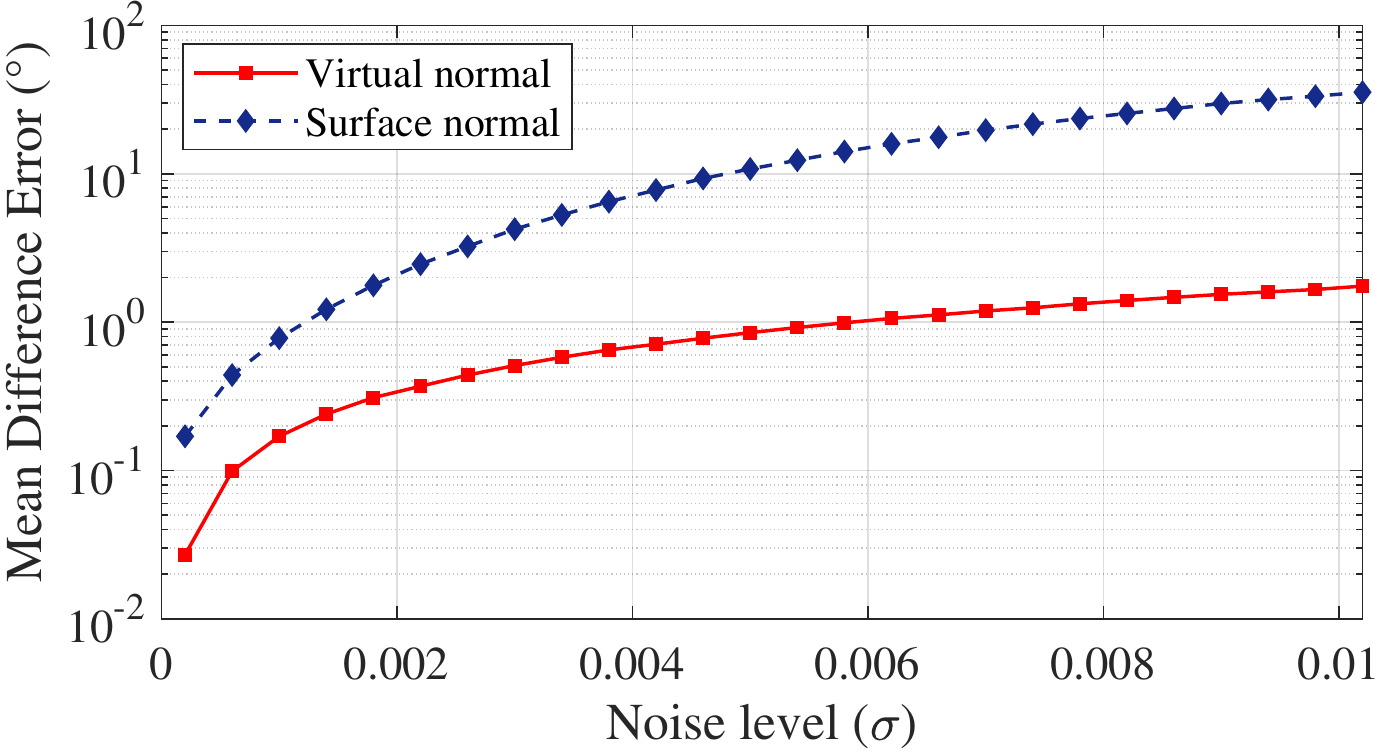}
	\label{fig:sn vn robustness results}
}
\caption{Robustness of virtual normal and surface normal against data noise. (a) The %
ideal surface and noisy surface. (b) The Mean Difference Error (Mean) is applied to evaluate the robustness of virtual normal and surface normal against different noise level. Our proposed virtual normal is more robust. } %
\label{fig:sn vn robustness}  %
\end{figure}

Furthermore, we conduct a simple experiment to verify the robustness of our proposed virtual normal against data noise. We create an unit sphere  and then add  gaussian noise  to simulate the ideal noise-free data and the real noisy data (see Fig.~\ref{fig:sn vn robustness sphere}). We then sample 100K groups points from the noisy surface and the ideal one to
compute
the virtual normal respectively, while 100K points are sampled to compute the surface normal as well. For the gaussian noise,
we use different deviations to simulate different noise levels by varying
deviation  $\sigma = [0.0002, ..., 0.01]$, and the mean being $\mu = 0$. The experimental results are illustrated in Fig.~\ref{fig:sn vn robustness results}. We can learn that our proposed virtual normal is much more robust to the data noise than the surface normal. Other local constraints are also sensitive to data noise.

Most `local'  geometric constraints, such as the surface normal, %
actually
enforcing
the first-order smoothness of the
surface
but are less useful for
helping
the  depth map prediction.
In contrast, the proposed VN establishes long-range relations in the 3D space.
Compared with pairwise CRFs, VN encodes triplet based relations, thus being of high
order.

\noindent\textbf{Virtual Normal Loss.}
We can sample a large number of triplets and compute corresponding VNs.
With the sampled VNs, we compute the divergence as
the Virtual Normal Loss (VNL):
\begin{equation}
    \def\bn{ { \boldsymbol n  } }
\label{eq:vnl}
    \ell_{VN} = \frac{1}{N} (
    \sum_{i=0}^{N}
    \| { \bn_{i}^{pred}} -{\bn_{i}^{gt}}\|_{1})
\end{equation}
where the $N$ is the number of valid sampling groups satisfying $\mathscr{R}_{1}, \mathscr{R}_{2}$. %
In experiments we have employed online hard example mining.

\noindent\textbf{Pixel-wise Depth Supervision.}
We also use a standard pixel-wise depth map loss.
We quantize  the real-valued depth and  formulate the depth prediction as a  classification problem instead of regression, and employ the cross-entropy loss.
In particular we  follow~\cite{cao2017estimating} to use the weighted cross-entropy loss (WCEL), with the weight being the information gain. See \cite{cao2017estimating} for details.
To obtain the accurate depth map and recover high-quality 3D information, we combine WCEL and VNL together to supervise the network output.
The overall loss is:
\begin{equation}
    \ell=\ell_{WCE} + \lambda \ell_{VN},
\label{eq: total loss}
\end{equation}
where $\lambda$ is a trade-off parameter, which is set to $5$ in all experiments to make the two terms roughly of the same scale.

Note that the above overall loss function is differentiable. The gradient of
the $ \ell_{VN}$ loss can be easily computed as Eq.~\eqref{eq:normal} and
Eq.~\eqref{eq:vnl} are both differentiable.

\section{Experiments}
In this section, we
conduct
several experiments to compare ours against  state-of-the-art methods.
We evaluate our methods on two datasets, NYUD-V2 and KITTI.

\subsection{Datasets}
\noindent\textbf{NYUD-V2.} The NYUD-V2 dataset consists of 464 different indoor scenes, which are further divided into 249 scenes for training and 215 for testing. We randomly sample 29K images from the training set to form NYUD-Large. Note that DORN uses the whole training set, which is significantly larger than that what we use. Apart from the whole dataset, there are officially annotated 1449 images (NYUD-Small), in which 795 images are split for training and others are for testing. In the ablation study, we use the NYUD-Small data.

\noindent\textbf{KITTI.} The KITTI dataset contains over 93K outdoor images and depth maps with the resolution around $1240\times374$. All images are captured on driving cars by stereo cameras and a Lidar. We test on 697 images from 29 scenes split by Eigen~\etal\cite{eigen2014depth}, validate on 888 images, and train on about 23488 images from the remaining 32 scenes.

\subsection{Implementation Details}

The pre-trained ResNeXt-101~\cite{xie2017aggregated} $(32 \times 4d)$ model on ImageNet~\cite{deng2009imagenet} is used  as our backbone model. A polynomial decaying method with the base learning rate 0.0001 and the power of $0.9$ is applied
for SGD.
The weight decay and the momentum are set to 0.0005 and 0.9 respectively.
Batch size is $8$ in our experiments.
The model is trained for $10$ epochs  on NYUD-Large and KITTI,
and is trained for $40$ epochs on NYUD-Small in the ablation study.
We perform the data augmentation on the training samples by the following methods. For NYUD-V2, the RGB image and the depth map are randomly resized with ratio $[1, 0.92, 0.86, 0.8, 0.75, 0.7, 0.67]$, randomly flipped in the horizon, and finally randomly cropped with the size $384 \times 384$ for NYUD-V2. The similar process is applied for KITTI but resizing with the ratio $[1, 1.1, 1.2, 1.3, 1.4, 1.5]$ and cropping with $384 \times 512$. Note that the depth map should be scaled with the corresponding resizing ratio.

\subsection{Evaluation Metrics}
We follow previous methods~\cite{laina2016deeper} to evaluate the performance of monocular depth prediction quantitatively based on following metrics: mean absolute relative error (rel),
mean $\log_{10}$ error ($ \log_{10}$), root mean squared error (rms) , root mean squared log error (rms (log)) and the accuracy under threshold ($\delta_{i} < 1.25^{i}, i=1, 2, 3$).

\subsection{Comparison with State-of-the-art}
In this section, we detail the comparison of our methods with state-of-the-art methods.

\begin{table}[!t]
\caption{Results on NYUD-V2. Our method outperforms other state-of-the-art methods over all evaluation metrics.}
\scalebox{0.75}{
\begin{tabular}{r |cccccc}
\toprule[1pt]
\multirow{2}{*}{Method} & \textbf{rel} & \textbf{log10} & \textbf{rms} & $\boldsymbol{\delta_{1}}$ & $\boldsymbol{\delta_{2}}$ & $\boldsymbol{\delta_{3}}$ \\
                        & \multicolumn{3}{c}{Lower is better}         & \multicolumn{3}{c}{Higher is better} \\ \hline
Saxena \etal.~\cite{saxena2009make3d}  & $0.349$  & -    & $1.214$   & $0.447$  & $0.745$  & $0.897$\\
Karsch \etal.~\cite{karsch2014depth}   & $0.349$  & $0.131$  & $1.21$   & -  & -  & -          \\
Liu \etal.~\cite{liu2014discrete}     & $0.335$  & $0.127$   & $1.06$  & -  & -  & -          \\
Ladicky \etal.~\cite{ladicky2014pulling}  & -   & -  & -    & $0.542$  & $0.829$  & $0.941$      \\
Li \etal.~\cite{li2015depth}   & $0.232$  & $0.094$  & $0.821$   & $0.621$ & $0.886$  & $0.968$      \\
Roy \etal.~\cite{roy2016monocular}   & $0.187$ & $0.078$  & $0.744$  & -  & -   & -          \\
Liu \etal.~\cite{liu2016learning}   & $0.213$  & $0.087$  & $0.759$  & $0.650$  & $0.906$ & $0.974$  \\
Wang \etal.~\cite{wang2015towards}  & $0.220$  & $0.094$  & $0.745$  & $0.605$  & $0.890$ & $0.970$ \\
Eigen \etal.~\cite{eigen2015predicting}  & $0.158$  & -     & $0.641$ & $0.769$  & $0.950$  & $0.988$ \\
Chakrabarti~\cite{chakrabarti2016depth}      & $0.149$  & - & $0.620$  & $0.806$  & $0.958$  & $0.987$ \\
Li \etal.~\cite{li2017two}   & $0.143$   & $0.063$  & $0.635$  & $0.788$  & $0.958$ & $0.991$    \\
Laina \etal.~\cite{laina2016deeper}   & $0.127$  & $0.055$    & $0.573$  & $0.811$   & $0.953$  & $0.988$      \\
DORN~\cite{fu2018deep}   & $0.115$  & $0.051$  & $0.509$   & $0.828$  & $0.965$  & $0.992$  \\ \hline \hline
Ours    & $\boldsymbol{0.108}$    & $\boldsymbol{0.048}$   & $\boldsymbol{0.416}$   & $\boldsymbol{0.875}$           & $\boldsymbol{0.976}$    & $\boldsymbol{0.994}$   \\ \toprule[1pt]
\end{tabular}\newline}
\label{table:errors cmp on NYUD-V2}
\end{table}

\noindent\textbf{NYUD-V2.} In this experiment, we compare with other state-of-the-art methods on the NYUD-V2 dataset. Table~\ref{table:errors cmp on NYUD-V2} demonstrates that our proposed method outperforms other state-of-the-art methods across all evaluation metrics significantly. Compare to DORN, we have improved the accuracy from $0.2\% $
to $18\%$ over all evaluation metrics that they report.

\begin{figure}[!bth]
\centering
\includegraphics[width=0.47\textwidth]{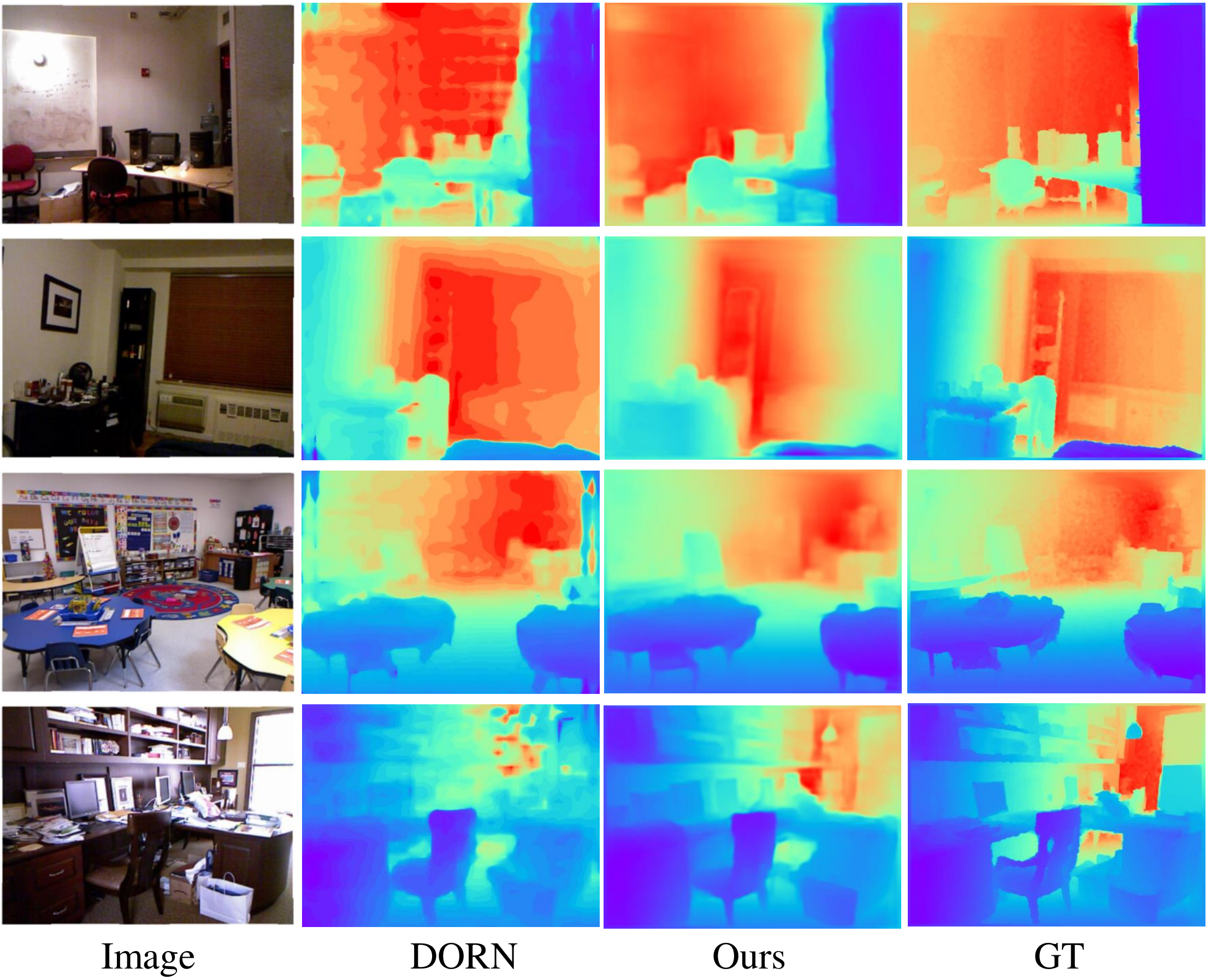}
\caption{Examples of predicted depth maps by our method and the state-of-the-art DORN on NYUD-V2. Color indicates the depth (red is far, purple is close). Our predicted depth maps have fewer errors in planes (see walls) and have high-quality details in complicated scenes (see the desk and shelf in the last row)}.
\label{fig:visual cmp}
\end{figure}

In addition to the quantitative comparison, we demonstrate some visual results between our method and the state-of-the-art DORN in Fig.~\ref{fig:visual cmp}. Clearly, the predicted depth by the proposed method is much more accurate. The plane of ours is much smoother and has fewer errors (see the wall regions colored with red in the 1st, 2nd, and 3rd row). Furthermore, the last row in Fig.~\ref{fig:visual cmp} manifests that our predicted depth is more accurate in the complicated scene. We have fewer errors in shelf and desk regions.

\noindent\textbf{KITTI.} In order to demonstrate that our proposed method can still reach the state-of-the-art performance on outdoor scenes, we test our method on the KITTI dataset. Results in Table~\ref{table:errors cmp on KITTI}
show
that our method has outperformed all other methods on all evaluation metrics except root mean square (rms) error.
The rms error is only
slightly behind that of DORN.
Note that for outdoor scenes, the rms (log) error, instead of rms, is usually the metric
of interest, in which ours is better.

\begin{table}[bth!]
\caption{Results on KITTI. Our method outperforms other methods over all evaluation metrics except rms.
}

\scalebox{0.75}{
\begin{tabular}{r |ccccccc}
\toprule[1pt]
\multirow{2}{*}{Method}     & $\boldsymbol{\delta_{1}}$ & $\boldsymbol{\delta_{2}}$ & $\boldsymbol{\delta_{3}}$         & \textbf{rel}   & \textbf{rms}   & \textbf{rms (log)} \\
 & \multicolumn{3}{c}{Higher is better} & \multicolumn{3}{c}{Lower is better} \\ \hline
Make3D \cite{saxena2009make3d}     & $0.601$  & $0.820$   & $0.926$  & $0.280$  & $8.734$    & $0.361$    \\
Eigen \etal. \cite{eigen2014depth} & $0.692$  & $0.899$  & $0.967$  & $0.190$ & $7.156$ & $0.270$    \\
Liu \etal. \cite{liu2016learning}  & $0.647$  & $0.882$  & $0.961$  & $0.114$  & $4.935$    & $0.206$    \\
Semi.\ \cite{kuznietsov2017semi} & $0.862$  & $0.960$  & $0.986$  & $0.113$ & $4.621$  & $0.189$    \\
Guo \etal. \cite{guo2018learning}  & $0.902$  & $0.969$  & $0.986$ & $0.090$ & $3.258$ & $0.168$    \\
DORN \cite{fu2018deep}  & $0.932$  & $0.984$  & $0.994$ & $0.072$ & $\boldsymbol{2.727}$  & $0.120$    \\\hline \hline
Ours & $\boldsymbol{0.938}$   & $\boldsymbol{0.990}$   & $\boldsymbol{0.998}$  & $\boldsymbol{0.072}$  & $3.258$      & $\boldsymbol{0.117}$    \\
\toprule[1pt]
\end{tabular}}
\label{table:errors cmp on KITTI}
\end{table}

\begin{figure*}[!ht]
\centering
\includegraphics[width=0.9\textwidth]{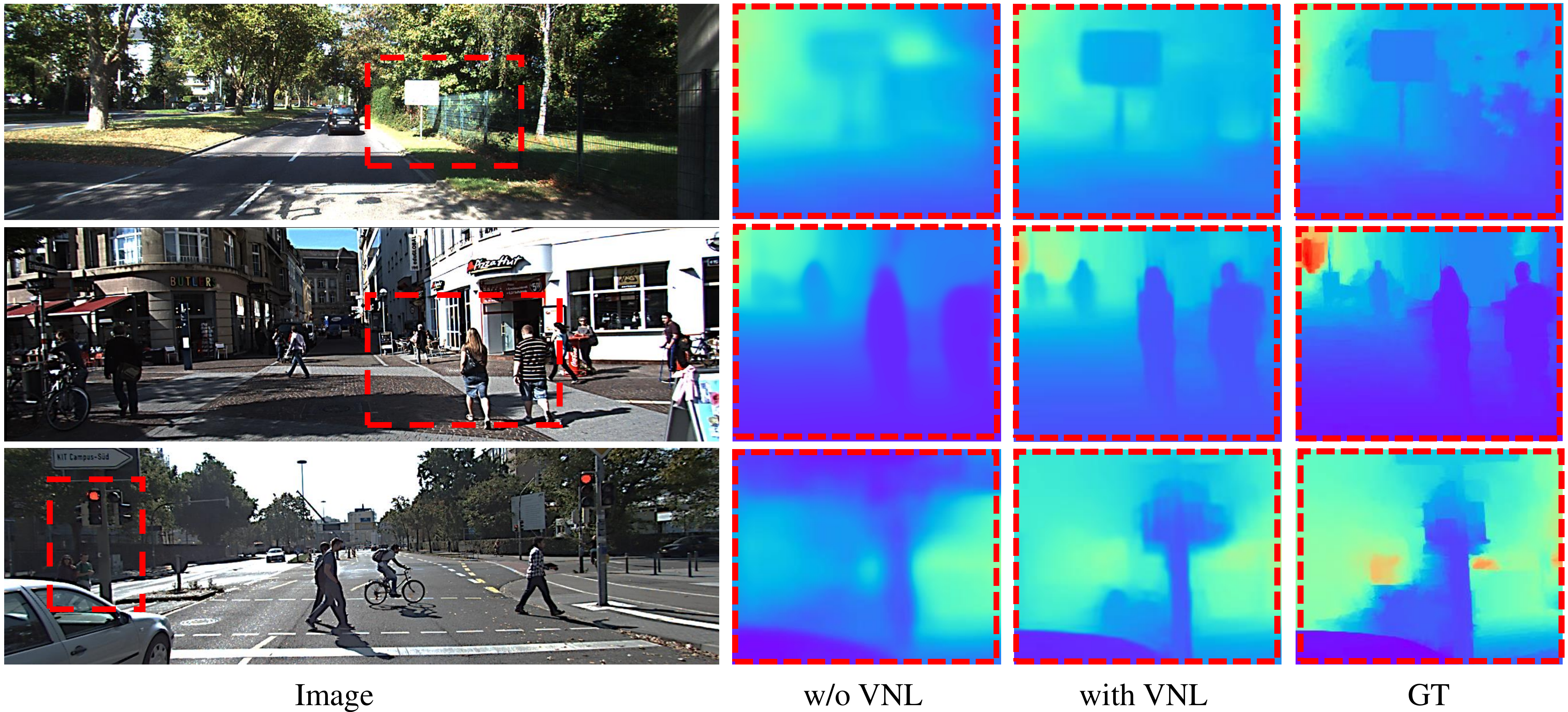}
\caption{Depth maps in the red dashed  boxes  with sign,
pedestrian and traffic lights
are zoomed in.
One can see that with the help of virtual normal,
predicted depth maps in these ambiguous regions are considerably more accurate.}
\label{fig:kitti dataset}
\end{figure*}

\subsection{Ablation Studies}
In this section, we conduct several ablation studies to analyze the details of our approach.

\noindent\textbf{Effectiveness of VNL.}
In this study, in order to prove the effectiveness of the proposed VNL
we compare it with two %
types
of pixel-wise depth map supervision, a pair-wise geometric supervision, and a high-order geometric supervision: 1) the ordinary cross-entropy loss (CEL); 2) the $L_{1}$ loss ($L_{1}$); 3) the surface normal loss (SNL); 4) the pair-wise geometric loss (PL). We reconstruct the point cloud from the depth map and further recover the surface normal from the point cloud. The angular discrepancy between the ground truth and recovered surface normal is defined as the surface normal loss, which is a high-order geometric supervision in 3D space. The pair-wise loss is the direction difference of two  vectors in 3D, which are established by randomly sampling paired points in ground-truth and predicted point cloud. The loss function of PL is as follow,
\begin{equation}
    \ell_{PL} = \frac{1}{N}\sum_{i=0}^{N}(1 - \frac{\overrightarrow{P_{Ai}^{\ast}P_{Bi}^{\ast}} \cdot \overrightarrow{P_{Ai}P_{Bi}}} {\left \| \overrightarrow{P_{Ai}^{\ast}P_{Bi}^{\ast}} \right \| \cdot \left \| \overrightarrow{P_{Ai}P_{Bi}} \right \|})
\end{equation}
where $(P_{A}^{\ast},P_{B}^{\ast})_{i} $ and $(P_{A}, P_{B})_{i}$ are paired points sampled from the ground truth and the predicted point cloud respectively. $N$ is the total number of  pairs.

We also employ the long-range restriction $\mathscr{R}_{2}$ for the paired points. Therefore, similar to VNL, PL can also be seen as a global geometric supervision in 3D space. The experimental results are reported  in Table.~\ref{table:Effectiveness of VNL and WCEL}. WCEL is the baseline for all following experiments.

\begin{table}[!tb]

\centering
\small
\caption{Illustration of the effectiveness of VNL.}
\begin{threeparttable}
\scalebox{0.7}{
\begin{tabular}{c|cccccc}
\toprule[1pt]
Metrics  & \textbf{rel}    & \textbf{log10} & \textbf{rms}   & $\boldsymbol{\delta_{1}}$ & $\boldsymbol{\delta_{2}}$ & $\boldsymbol{\delta_{3}}$        \\ \hline \hline
\multicolumn{6}{c}{Pixel-wise Depth Supervision}               \\ \hline \hline
CEL      & $0.1456$ & $0.061$ & $0.617$ & $0.8087$ & $0.9559$ & $0.9862$ \\
WCEL     & $0.1427$ & $0.060$ & $0.511$ & $0.8117$ & $0.9611$ & $0.9895$ \\
WCEL+L1  & $0.1429$ & $0.061$ & $0.626$ & $0.8098$ & $0.9539$ & $0.9858$ \\ \hline \hline
\multicolumn{6}{c}{Pixel-wise Depth Supervision + Geometric Supervision}  \\ \hline \hline
WCEL+PL\tnote{\ddag}     & $0.1380$ & $0.059$ & $0.504$ & $0.8212$ & $0.9643 $ & $0.9913$ \\
WCEL+PL+VNL & $0.1341$ & $0.056$ & $0.485$ & $\boldsymbol{0.8336}$ & $\boldsymbol{0.9671}$ & $0.9913$ \\
WCEL+SNL\tnote{\dag} & $0.1406$ & $0.059$ & $0.599$ & $0.8209$ & $0.9602$ & $0.9886$ \\
WCEL+VNL\tnote{\ddag}~ (Ours) & $\boldsymbol{0.1337}$ & $\boldsymbol{0.056}$ & $\boldsymbol{0.480}$ & $0.8323$ & $0.9669$ & $\boldsymbol{0.9920}$ \\
\toprule[1pt]
\end{tabular}}
\begin{tablenotes}
\footnotesize
\item[\dag]`Local' geometric supervision in 3D.
\item[\ddag]`Global' geometric supervision in 3D.
\end{tablenotes}
\end{threeparttable}
\vspace{0.5em}

\label{table:Effectiveness of VNL and WCEL}

\end{table}

Firstly, we analyze the effect of pixel-wise depth supervision for prediction performance. As WCE employs an %
weight in the CE loss,
its performance is slightly better than that of CEL. However, when we enforce two pixel-wise supervision (WCEL+L1) on the depth map, the performance cannot improve any more. Thus using two  pixel-wise loss terms does not help.

Secondly, we analyze the effectiveness of the supplementary 3D geometric constraint (PL, SNL, VNL). Compared with the baseline (WCEL), three supplementary 3D geometric constraints can promote the network performance  with varying degrees. Our proposed VNL combining with WCEL has the best performance, which has improved the baseline performance by up to $8$\%.

Thirdly, we analyze the difference of three geometric constraints. As SNL can only %
exploit
geometric relations of homogeneous local regions, its performance
is the lowest among
the three constraints over all evaluation metrics. Compared with SNL, since PL constrains the global geometric relations, its performance is
clearly
better. However, the performance of WCEL+PL
is not as good as
our proposed WCEL+VNL.
When we further add our VNL  on top of  WCEL+PL, the precision can further
be slightly improved
and is comparable to WCEL+VNL. Therefore, although PL is a global geometric constraint in 3D, the pair-wise constraint cannot encode as strong geometry information as our proposed VNL.

At last, in order to further demonstrate the effectiveness of VNL, we analyze the results of network trained with and without VNL supervision on the  KITTI dataset. The visual comparison is shown  in Fig.~\ref{fig:kitti dataset}. One can see that VNL can
improve the performance of the network in ambiguous regions. For example, the sign (1st row), the distant pedestrian (2nd row), and traffic light in the last row of the figure can demonstrate the effectiveness of
the proposed VNL.
In conclusion, the geometric constraints in the 3D space can significantly
boost
the network performance. Moreover, the global and high-order constraints can enforce stronger supervision than the `local' and pair-wise ones in 3D space.

\begin{figure}[!ht]
\centering
\includegraphics[width=0.37\textwidth]{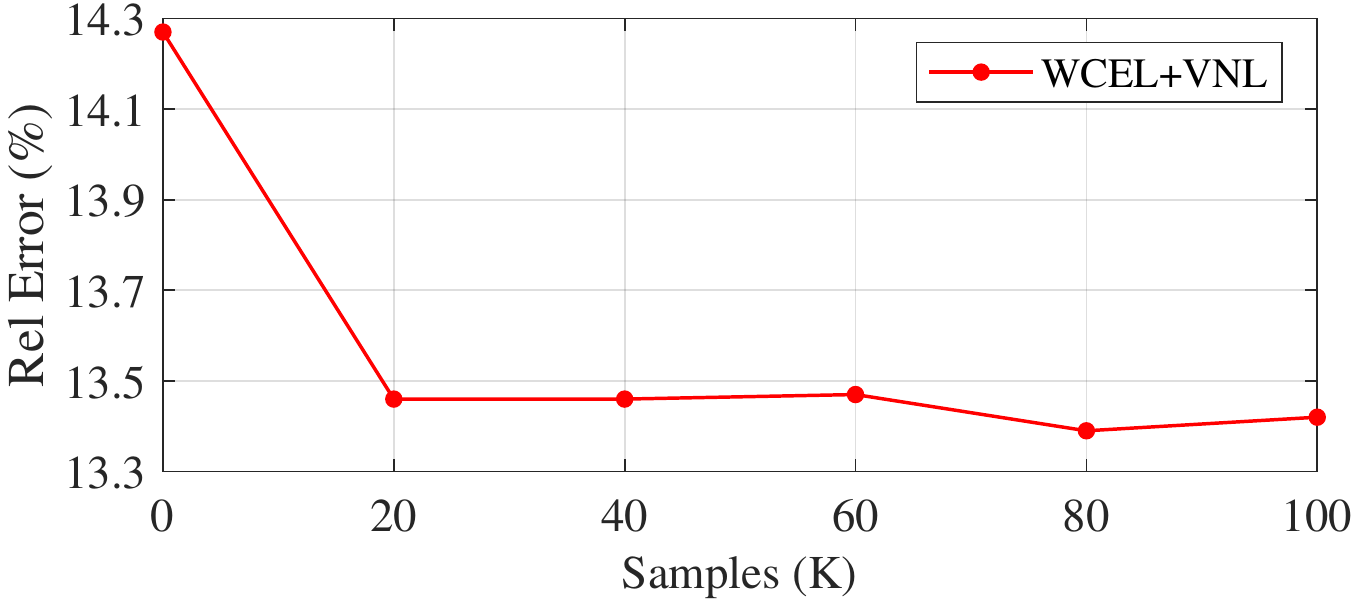}

\caption{Illustration of the impact of the samples size. The more samples will promote the performance.}
\label{fig:VNL_PL_Samples}
\end{figure}

\noindent\textbf{Impact of the Amount of Samples.}
Previously, we have proved the effectiveness of VNL. Here the impact of the size of samples for VNL
is
discussed. We sample six different sizes of point groups, 0K, 20K, 40K, 60K, and 80K and 100K, to establish VNL. `0K' means that the model is trained without VNL supervision. The rel error is reported for evaluation. Fig.~\ref{fig:VNL_PL_Samples} demonstrates that `rel'  slumps  by $5.6$\% with 20K point groups to establish VNL. However, it only drops slightly when the samples for VNL increase from 20K to 100K. Therefore, the performance saturates   with more samples,
when samples reach a certain number in that the diversity of samples is enough to construct the global geometric constraint.
\noindent\textbf{Lightweight Backbone Network.}
We train the network with the MobileNetV2 backbone to evaluate the effectiveness of the proposed geometric constraint on the light network. We train it on the NYUD-Large for $10$ epochs. Results in Table~\ref{table:errors cmp with MobileNetV2} show that the
proposed
VNL can
improve
the performance by $1\%$ - $8\%$. Comparing with previous state-of-the-art methods, we have improved the accuracy by around  $29\%$ over all evaluation metrics and achieved a better trade-off between parameters and the accuracy.

\begin{table}[!ht]
\caption{Performance on NYUD-V2 with MobileNetV2 backbone.
$^\dag${Trained without VN.} ~
$^\ddag${Trained with VN.}}
\centering
\small
\begin{threeparttable}
\scalebox{0.95}{
\begin{tabular}{c|cc|cc}
\toprule[1pt]
Metrics  &CReaM \cite{spek2018cream}  &RF-LW\cite{nekrasov2018real}  &Ours-B{$^\dag$}  &Ours-VN{$^\ddag$}\\ \hline
$\boldsymbol{\delta_{1}}$     & $0.704$    & $0.790$  &$0.814$  &$\boldsymbol{0.829}$   \\
$\boldsymbol{\delta_{2}}$     & $0.917$    & $0.955$  &$0.947$  &$\boldsymbol{0.956}$  \\
$\boldsymbol{\delta_{3}}$     & $0.977$    & $\boldsymbol{0.990}$ &$0.972$  & $0.980$  \\
\textbf{rel}                  & $0.190$    & $0.149$  &$0.144$   &$\boldsymbol{0.134}$  \\
\textbf{rms}                  & $0.687$    & $0.565$  &$0.502$   &$\boldsymbol{0.485}$  \\
\textbf{rms (log)}            & $0.251$    & $0.205$  &$0.201$   &$\boldsymbol{0.185}$\\ \hline
\textbf{params}               &$\boldsymbol{1.5}$\textbf{M}  &$3.0$M  &$2.7$M  &$2.7$M  \\
\toprule[1pt]
\end{tabular}}
\end{threeparttable}
\label{table:errors cmp with MobileNetV2}
\end{table}

\subsection{Recovering 3D Features from Estimated  Depth}
We have argued that, with geometric constraints in the 3D space, the network can achieve more accurate depth and also obtain higher-quality 3D information.
Here we show the recovered  3D point cloud and the  surface normal to support this.

\begin{figure}[!bth]
\centering
\includegraphics[width=0.45\textwidth]{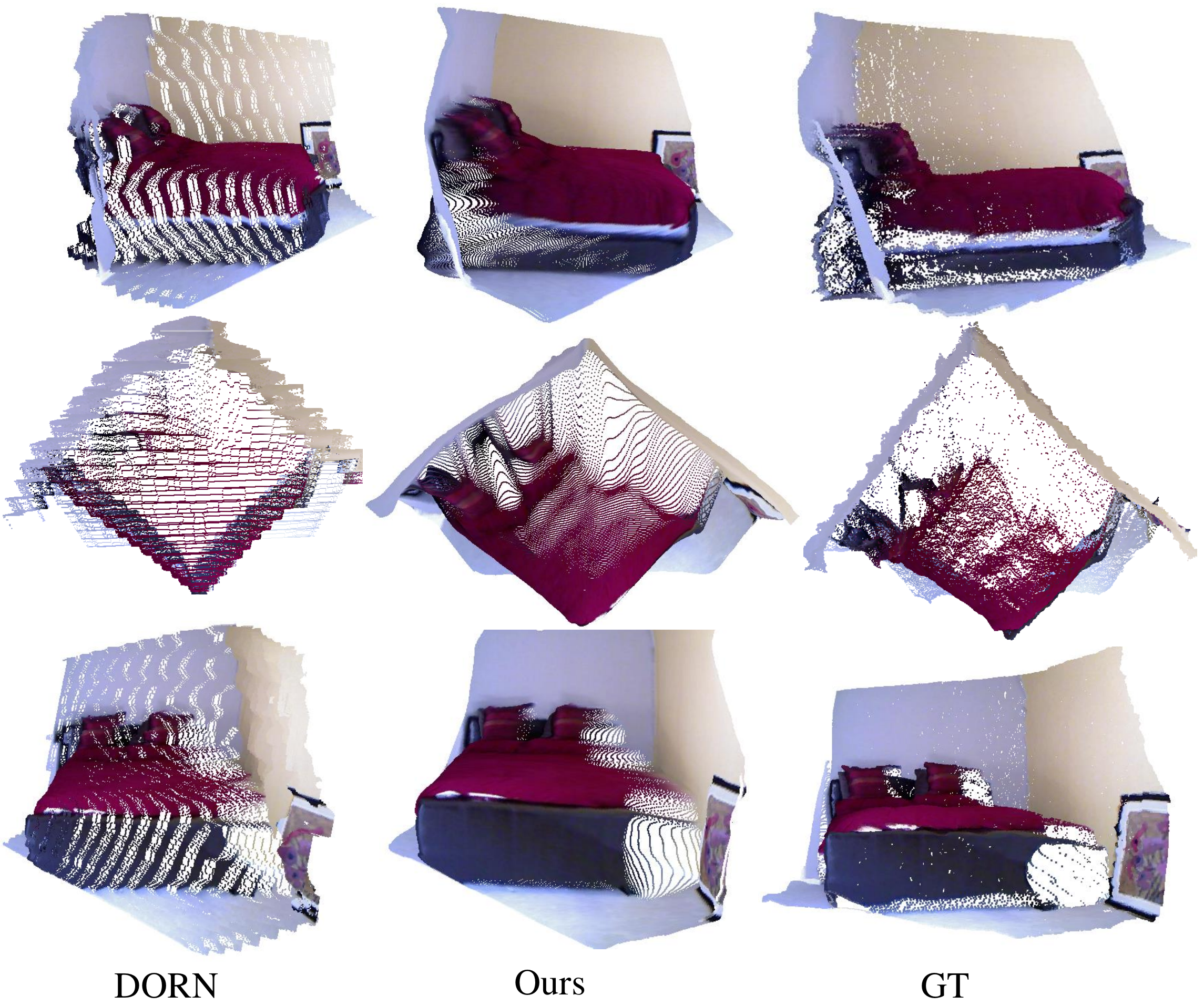}
\caption{Comparison of reconstructed point clouds from estimated depth maps between DORN
\cite{fu2018deep} and ours. %
We can see that our point cloud results contain less noise and are
closer to groud-truth
than that of DORN.}
\label{fig:pcd cmp}
\end{figure}

\noindent\textbf{3D Point Cloud.}
Firstly, we compare the reconstructed 3D point cloud from our predicted depth and that of DORN. Fig.~\ref{fig:pcd cmp}
demonstrate
that the overall quality of ours outperforms theirs   significantly. Although our predicted depth is only slightly better than theirs  on evaluation metrics, the reconstructed wall (see the 2nd row in ~\ref{fig:pcd cmp}) of ours is much flatter and has fewer errors. The shape of the bed is more similar to the ground truth. From the bird view, it is hard to recognize the bed shape of their results. The point cloud in Fig.~\ref{fig:pcd normal dorn ours} also leads to  a similar conclusion.

\noindent\textbf{Surface Normal.}
Lastly, we compare the calculated surface normal with previous state-of-the-art methods and demonstrate the quantitative results in Table~\ref{table:surface normals cmp}. The ground truth is obtained as described in~\cite{eigen2015predicting}. We first compare our geometrically calculated results with DCNN-based optimization methods. Although we do not optimize a sub-model to achieve the surface normal, our results can outperform most of the previous methods and even are the best on $\boldsymbol{30}\degree$ metric.

\begin{table}[!ht]
\caption{Evaluation of the surface normal on NYUD-V2.}
\centering
\small
\begin{threeparttable}
\scalebox{0.9}{
\begin{tabular}{r|cccccc}
\toprule[1pt]
\multirow{2}{*}{Method}     & $\textbf{Mean}$ & $\textbf{Median}$  & $\boldsymbol{11.2\degree}$ & $\boldsymbol{22.5\degree}$  & $\boldsymbol{30\degree}$ \\
           & \multicolumn{2}{c}{Lower is better} & \multicolumn{3}{c}{Higher is better} \\ \hline \hline
\multicolumn{6}{c}{Predicted Surface Normal from the Network}               \\ \hline \hline
3DP \cite{fouhey2013data}    & $33.0$      & $28.3$    & $18.8$    & $40.7$    & $52.4$    \\
Ladicky \etal.\ \cite{zeisl2014discriminatively}   & $35.5$   & $25.5$   & $24.0$  & $45.6$  & $55.9$    \\
Fouhey \etal.\ \cite{fouhey2014unfolding}    & $35.2$  & $17.9$   & $40.5$  & $54.1$    & $58.9$    \\
Wang \etal.\ \cite{wang2015designing}      & $28.8$   & $17.9$  & $35.2$  & $57.1$    & $65.5$    \\
Eigen \etal.\ \cite{eigen2015predicting}     & $\boldsymbol{23.7}$   & $\boldsymbol{15.5}$  & $\boldsymbol{39.2}$  & $\boldsymbol{62.0}$    & $71.1$    \\\hline \hline
\multicolumn{6}{c}{Calculated Surface Normal from the Point cloud}               \\ \hline \hline
GT-GeoNet\tnote{\dag} ~\cite{qi2018geonet}        & $36.8$    & $32.1$   & $15.0$    & $34.5$    & $46.7$ \\
DORN\tnote{\ddag} ~\cite{fu2018deep}   & $36.6$    & $31.1$      & $15.7$    & $36.5$    & $49.4$ \\
Ours             & $24.6$     & $17.9$     & $34.1$    & $60.7$    & $\boldsymbol{71.7}$    \\
\toprule[1pt]
\end{tabular}}

\begin{tablenotes}
\footnotesize
\item[\dag]Cited from the original paper.
\item[\ddag]Using authors' released models.
\end{tablenotes}
\end{threeparttable}
\label{table:surface normals cmp}
\end{table}

\begin{figure}[!ht]
\centering
\includegraphics[width=0.47\textwidth]{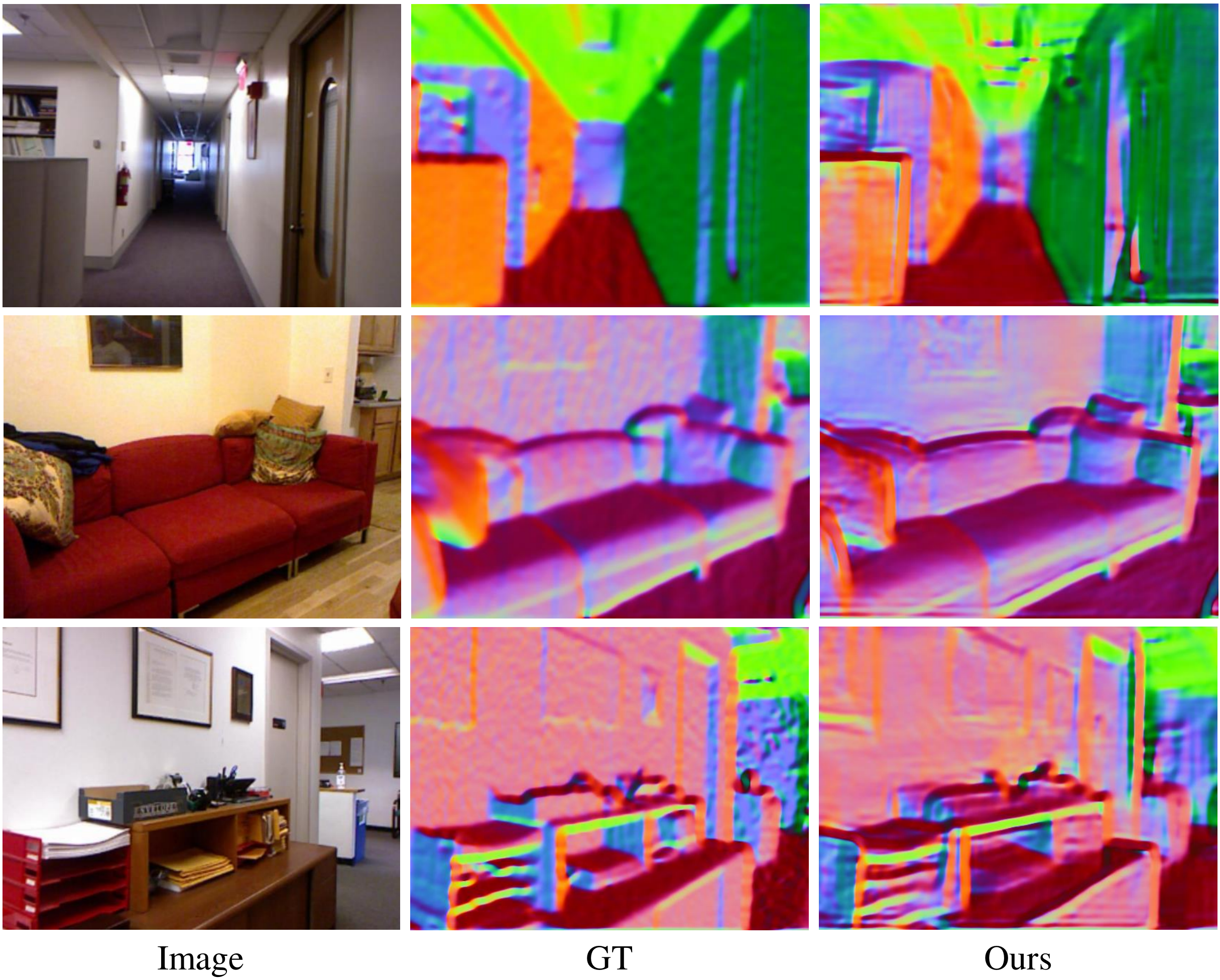}

\caption{Recovered surface normal from 3D point cloud. According to the visual effect, the surface normal is in high-quality in planes (1st row) and the complicated curved surface (2nd and last row).}
\label{fig:normals}
\end{figure}

Furthermore, we compare %
the surface normals  directly  computed  from the reconstructed point cloud with that of DORN ~\cite{fu2018deep} and GeoNet ~\cite{qi2018geonet}. Note that we run the released code and model of DORN to obtain depth maps and then calculate  surface normals from the depth, while the evaluation of GeoNet is cited from the original paper. In Table~\ref{table:surface normals cmp}, we can see that, with high-order geometric supervision, our method outperforms DORN and GeoNet by a large margin, and even is close to Eigen method which trains to output normals. It suggests that our method can lead the model to learn the shape from images.

Apart from the quantitative comparison, the visual effect is %
shown
in Fig.~\ref{fig:normals}, %
demonstrating
that our directly calculated surface normals are not only accurate in planes (the 1st row), but also are of higher quality in regions with sophisticated curved surface (the 2nd and last row).

 \section{Conclusion}
In this paper, we have proposed to construct a {\it long-range}  geometric constraint (VNL) in the 3D space for monocular depth prediction. In contrast to previous methods with only pixel-wise depth supervision in 2D space, our method can not only obtain the accurate depth maps but also recover high-quality 3D features, such as the point cloud and the surface normal, eliminating necessities to optimize a new sub-model. Compared with other 3D constrains, our proposed VNL is more robust to noise and can encode strong global constraints. Experimental results on NYUD-V2 and KITTI have proved the effectiveness of our method and the state-of-the-art performance.

In particular, to demonstrate that  our method is able to produce sensible local shapes,
the normals directly derived  from the estimated depth of our method outperform many other recent depth estimation methods and are close to
that of those trained to output normals.
We hope that our method  provides  a useful tool and stimulates  insight into predicting not only depth but also shape from monocular images.

\section{Appendix}
\subsection{Model}
\begin{figure}[!ht]
\centering
\includegraphics[width=0.45\textwidth]{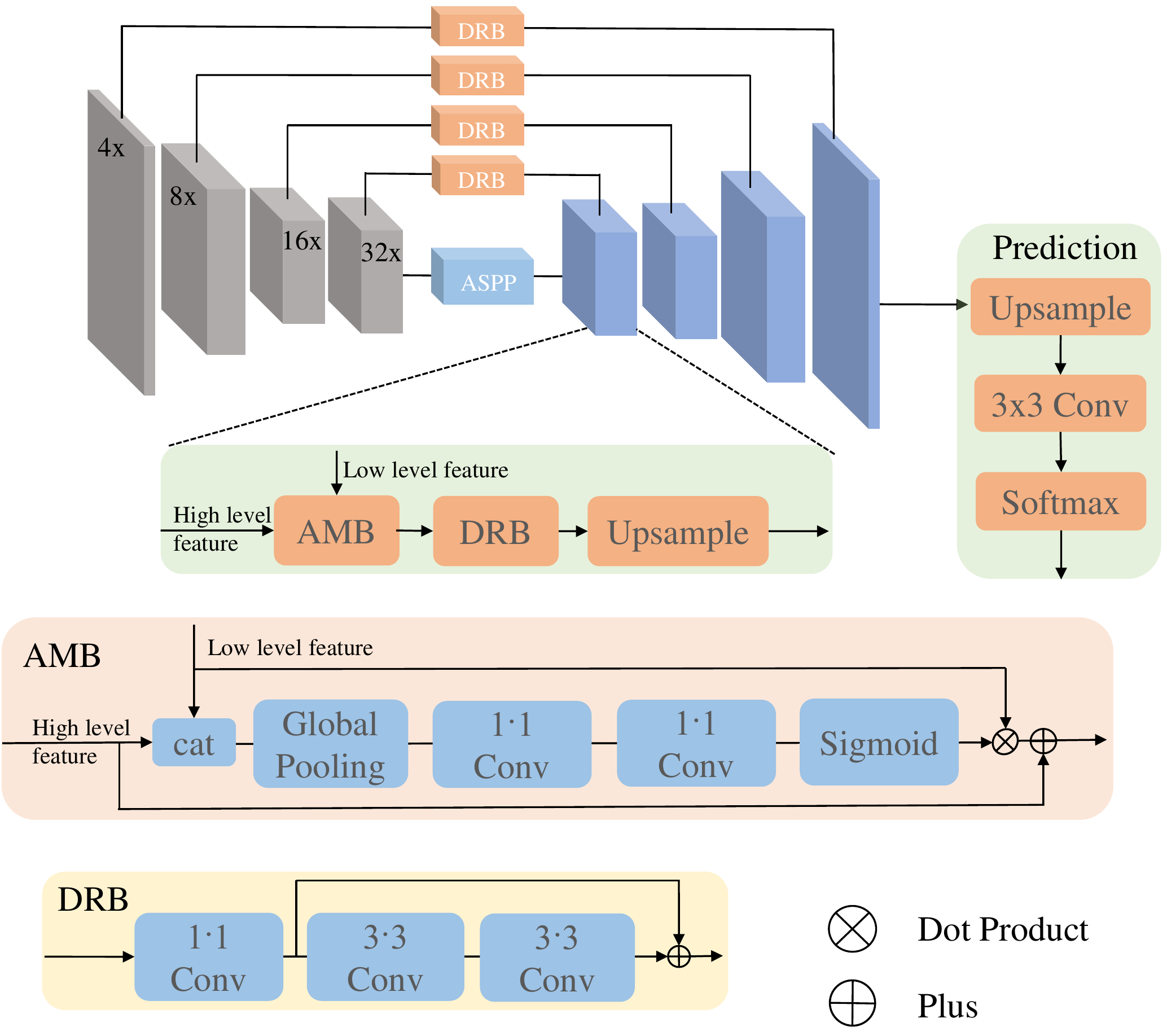}
\caption{Model architecture. The encoder-decoder network has four flip connections to merge low-level features.}
\label{fig:model architecture}
\end{figure}

An overview architecture of our model is illustrated in Fig.\ref{fig:model architecture}. The network is mainly composed of two parts, an encoder to establish features in different levels from $I_{in}$, and a decoder to reconstruct the depth map. Inspired by~\cite{li2018deep}, the decoder is composed of several adaptive merging blocks (AMB) to fuse features from different levels and dilated residual blocks (DRB) to transform features. In order to improve the receptive field of the decoder, we set the dilation rates of all $3\times3$ convolutions in DRB to 2 and insert an Astrous Spatial Pyramid Pooling (ASPP) module (dilation rate: 2, 4, 8)~\cite{chen2018encoder} between the encoder and the decoder. Furthermore, we establish 4 flip connections from different levels of encoder blocks to the decoder to merge more low-level features. The AMB will learn a merging parameter for adaptive merging. Apart from features from the highest level with 512 channels, other flips' features dimension are 256. At last, a prediction module, a $3 \times 3$ convolution and a softmax, is applied to transfer the features dimensions from 256 channels to 150 depth bins.

In the lightweight backbone network experiment, the backbone is replaced with MobileNetV2. In order to further reduce parameters, the dimensions of four flip connections are reduced to $(128, 64, 64, 64)$. In the prediction module, the features are transferred from 64 channels to 60 depth bins.

\subsection{Predicted Depth and Surface Normal}
\begin{figure*}[!tb]
\centering
\includegraphics[width=.8\textwidth]{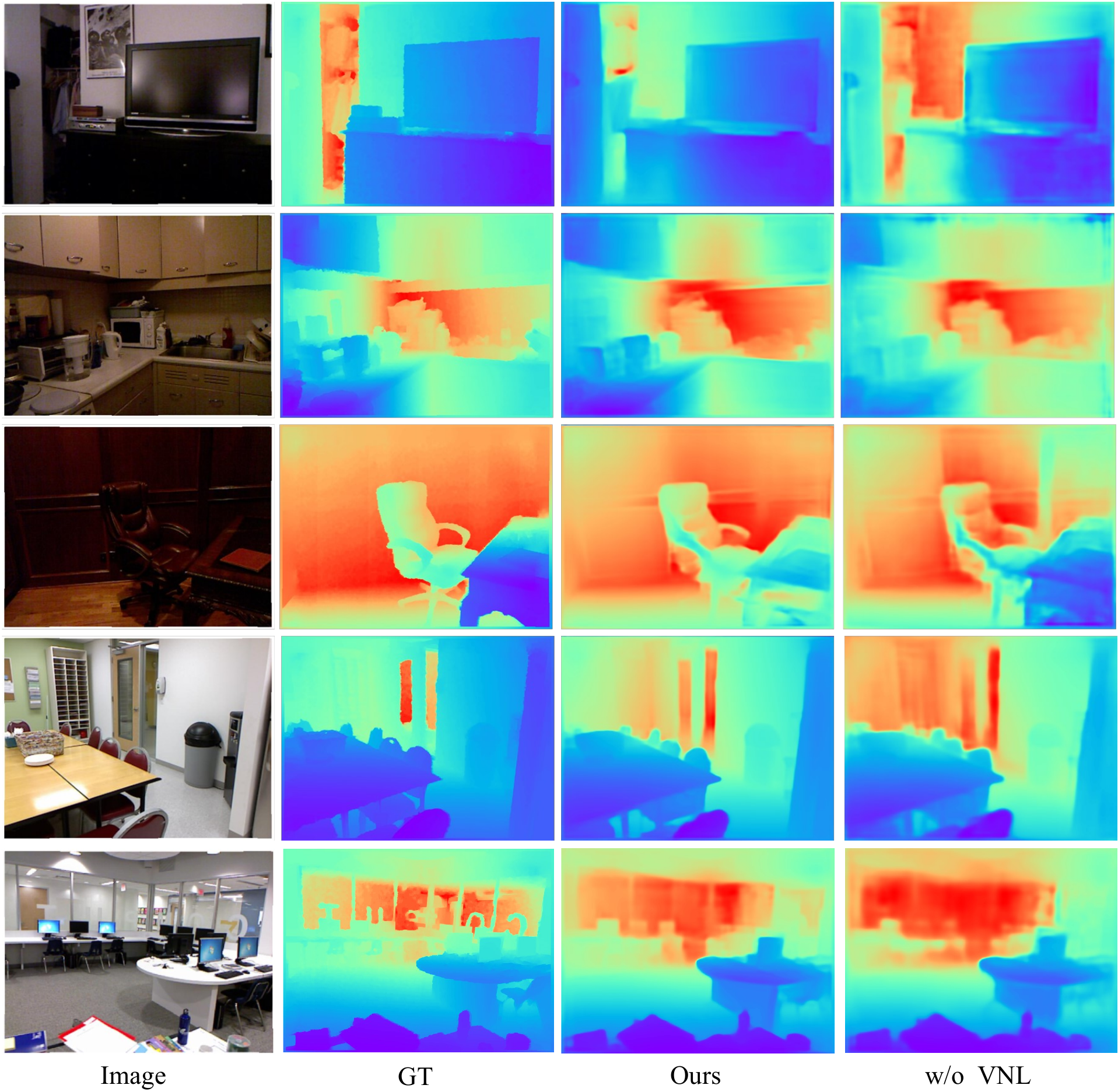}
\caption{Samples of the predicted depth on NYUD-V2. By adding the virtual normal constraints, our proposed model can produce more accurate and smooth depth map.}
\label{fig:nyu-depth-sup}
\end{figure*}

\begin{figure*}[!ht]
\centering
\includegraphics[width=0.758\textwidth]{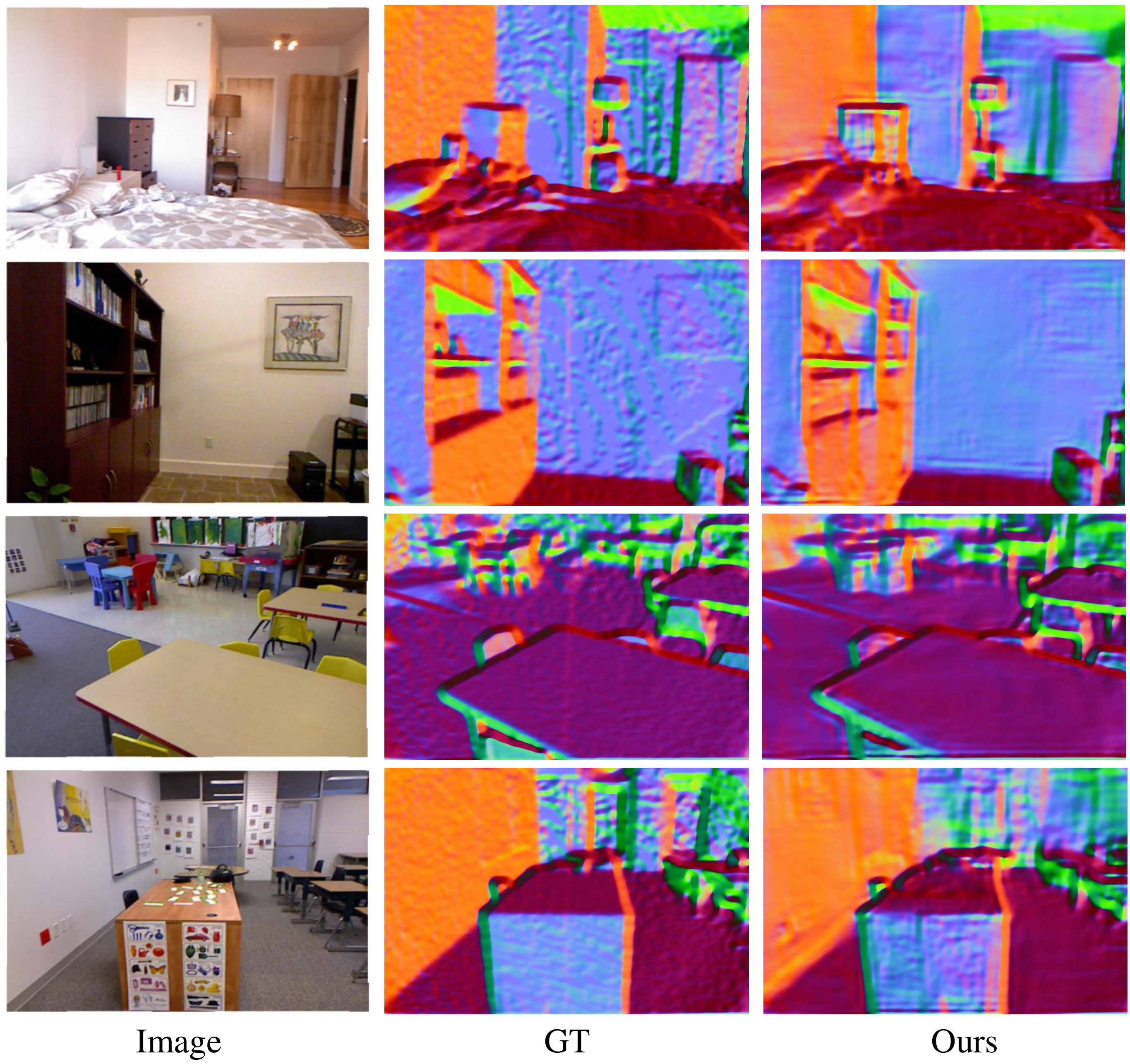}
\caption{Samples of the recovered surface normals on NYUD-V2. One can see that we can recover surface normals from the reconstructed point cloud with high quality .}
\label{fig:nyu-normal-sup}
\vspace{-1em}
\end{figure*}

\begin{figure*}[!ht]
\centering
\includegraphics[width=0.85\textwidth]{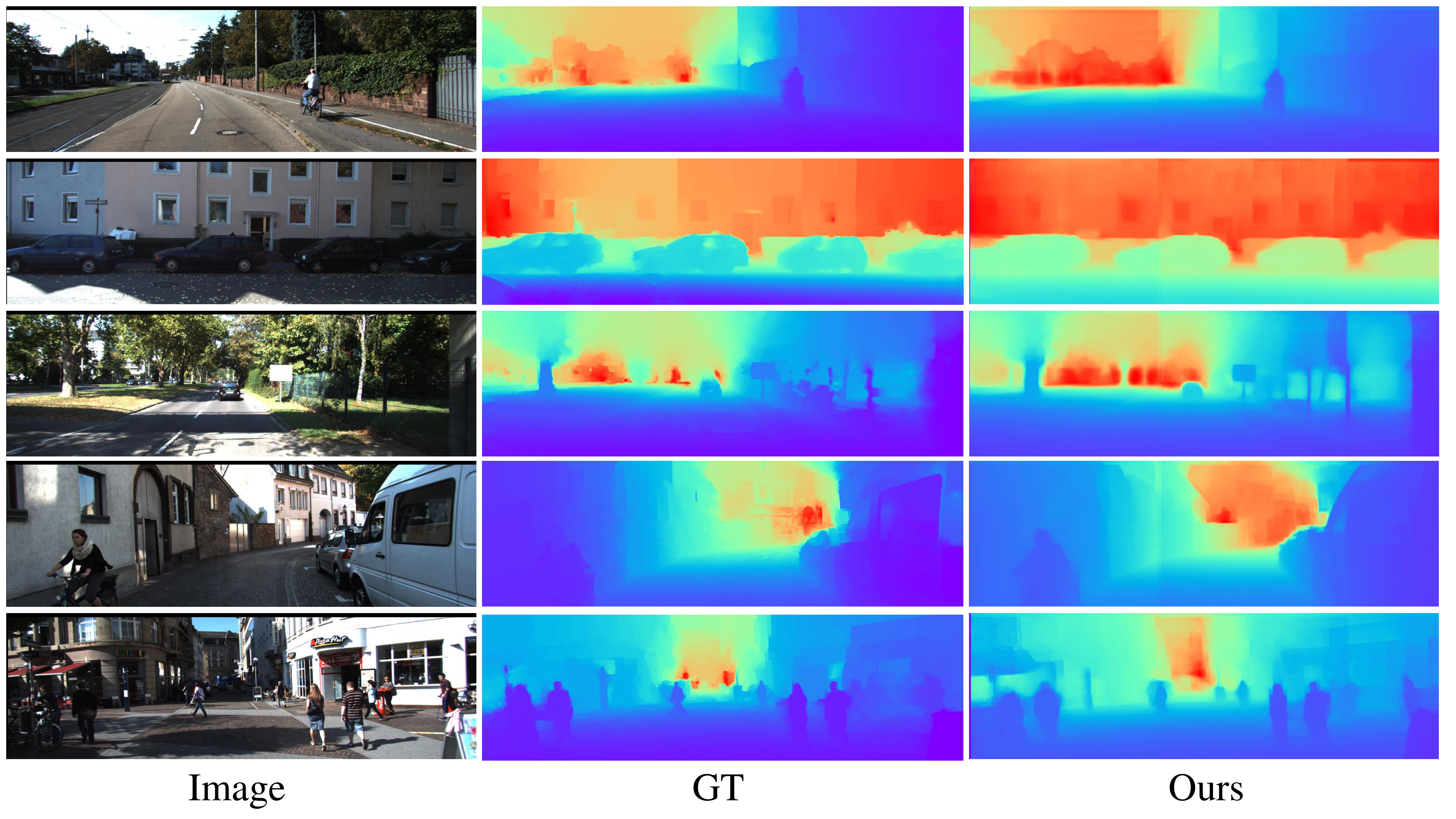}
\caption{Samples of the predicted depth on KITTI. The high quality results  show the effectiveness of our methods.  }
\label{fig:kitti-depth-sup}
\end{figure*}

We provide more predicted depth maps and recovered surface normals on KITTI and NYUD-V2 dataset. Depth maps are illustrated in Fig.~\ref{fig:nyu-depth-sup},  and Fig.~\ref{fig:kitti-depth-sup}, the recovered surface normals are demonstrated in Fig.~\ref{fig:nyu-normal-sup}. %

\subsection{3D point cloud}
\begin{figure*}
\centering    %
\subfloat[] %
{
	\includegraphics[width=0.8\textwidth]{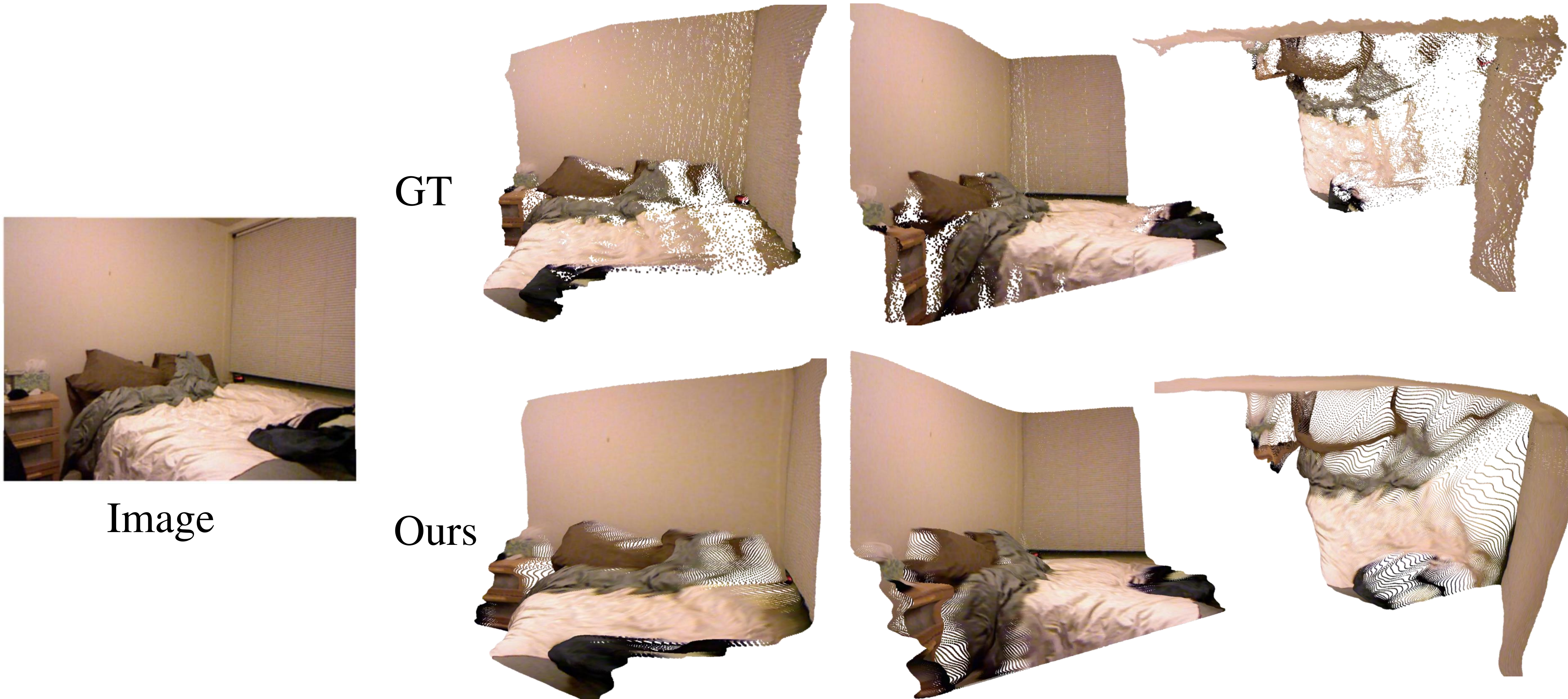}
}

%
%

\subfloat[] %
{
	\includegraphics[width=0.8\textwidth]{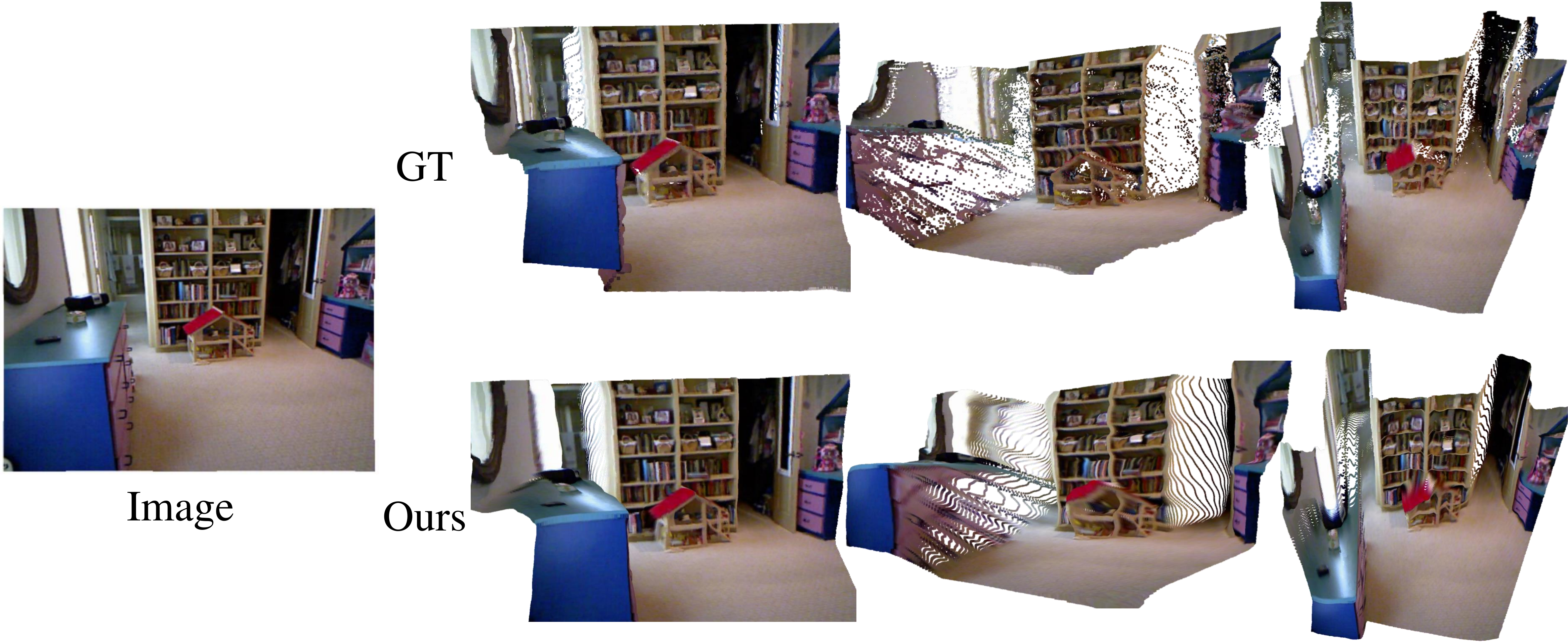}
	\label{fig:sn vn robustness results}
}
%

\subfloat[] %
{
	\includegraphics[width=0.8\textwidth]{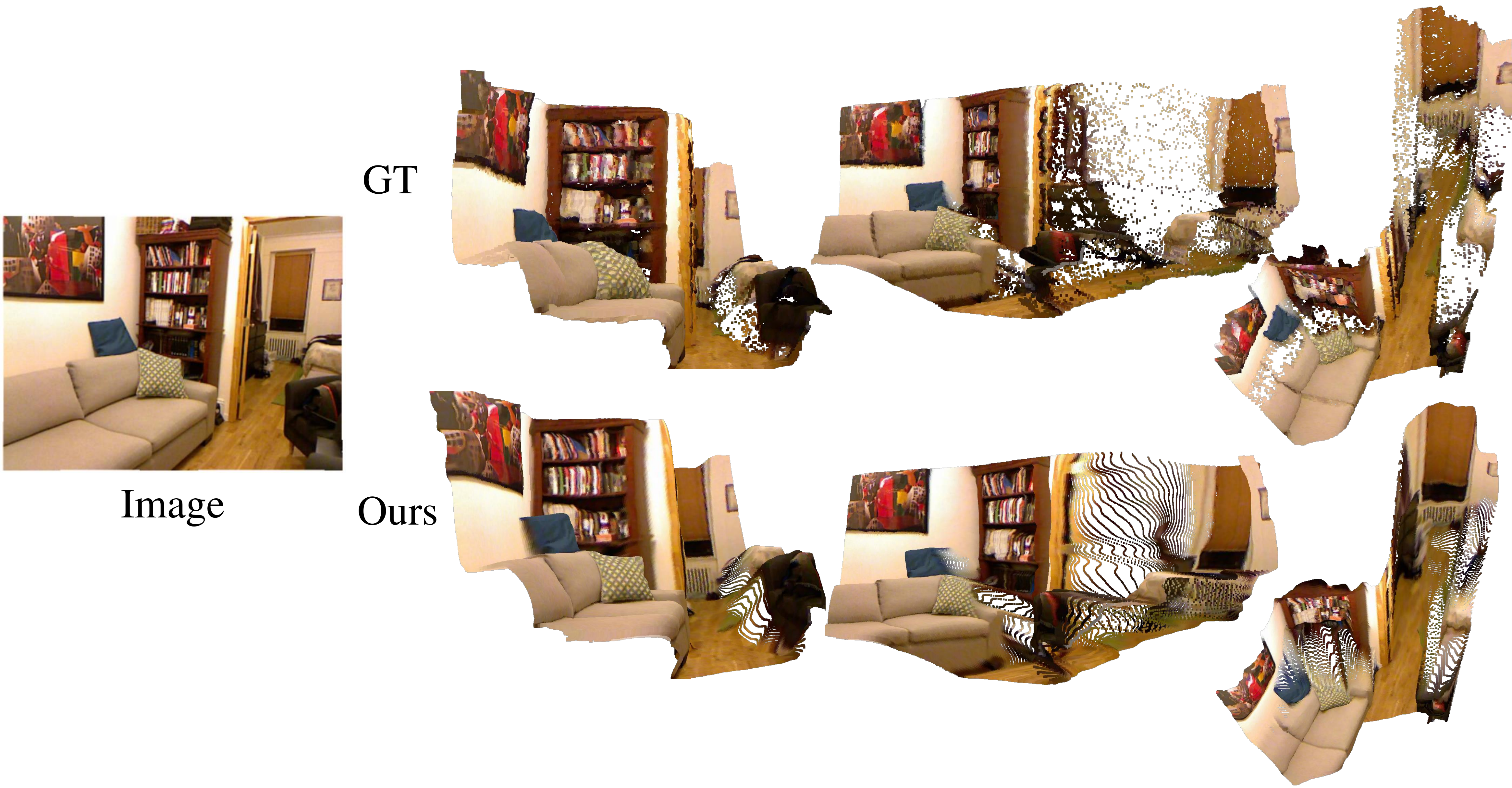}
	\label{fig:sn vn robustness results}
}

\caption{Reconstructed point clouds. Three scenes are randomly selected from NYUD-V2. For the reconstructed point cloud of each scene, 3 views are selected to demonstrate the point cloud. (a) Scene 1; (b) Scene 2; (c) Scene 3. } %
\label{fig:pcd_nyudv2}  %

\end{figure*}

\begin{figure*}
\centering    %
\subfloat[] %
{
	\includegraphics[width=0.8\textwidth]{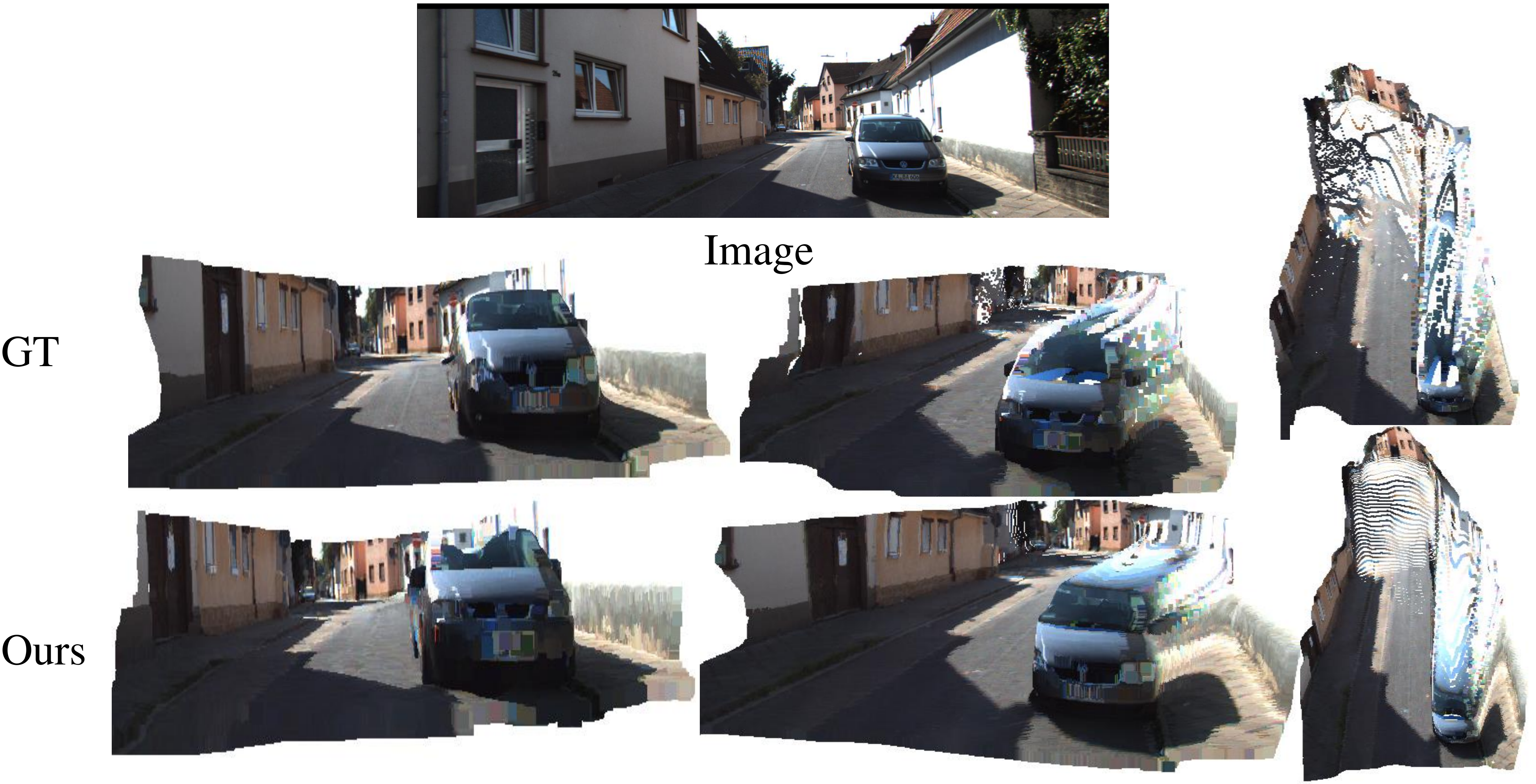}
}
\\
%

\subfloat[] %
{
	\includegraphics[width=0.8\textwidth]{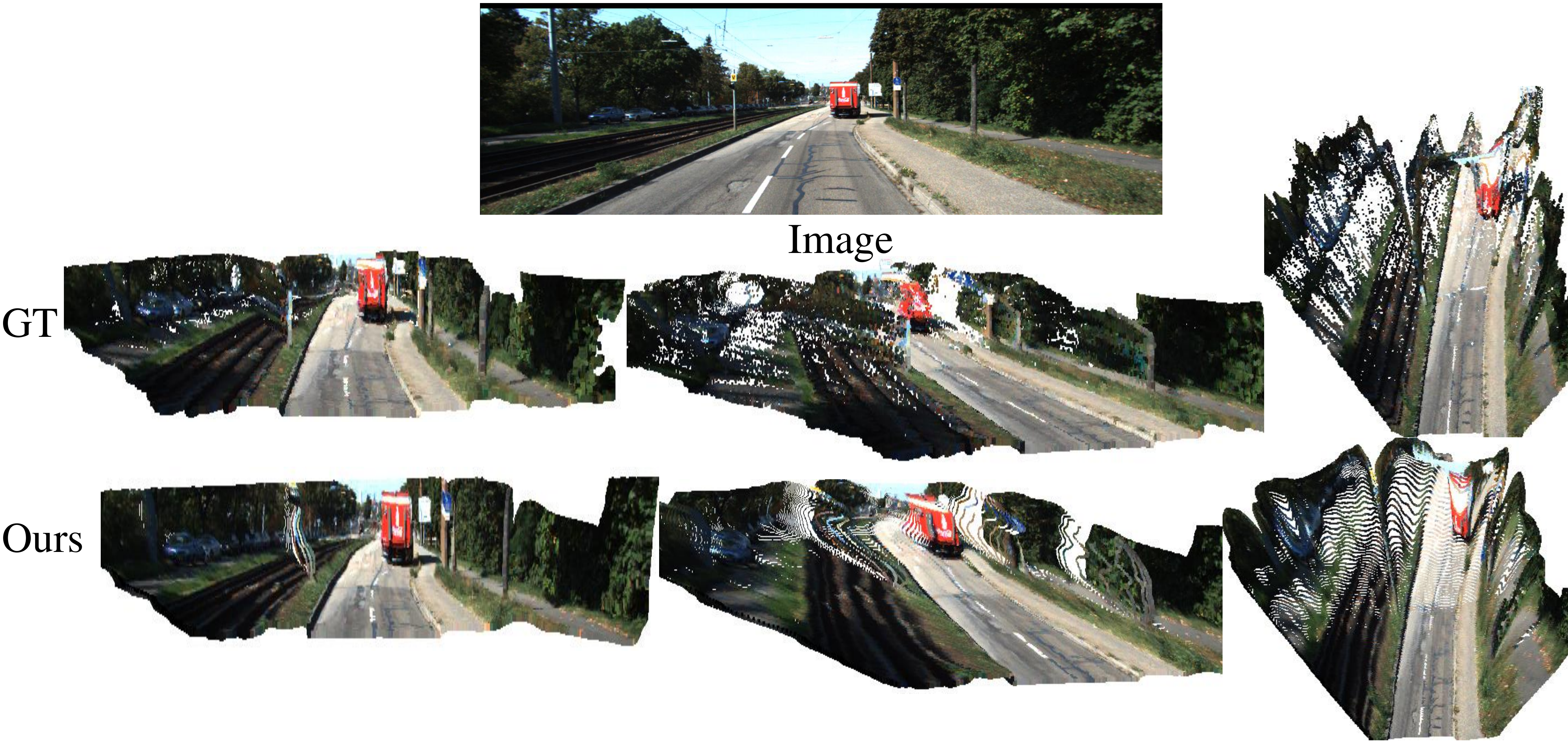}
}\\
\subfloat[] %
{
	\includegraphics[width=0.8\textwidth]{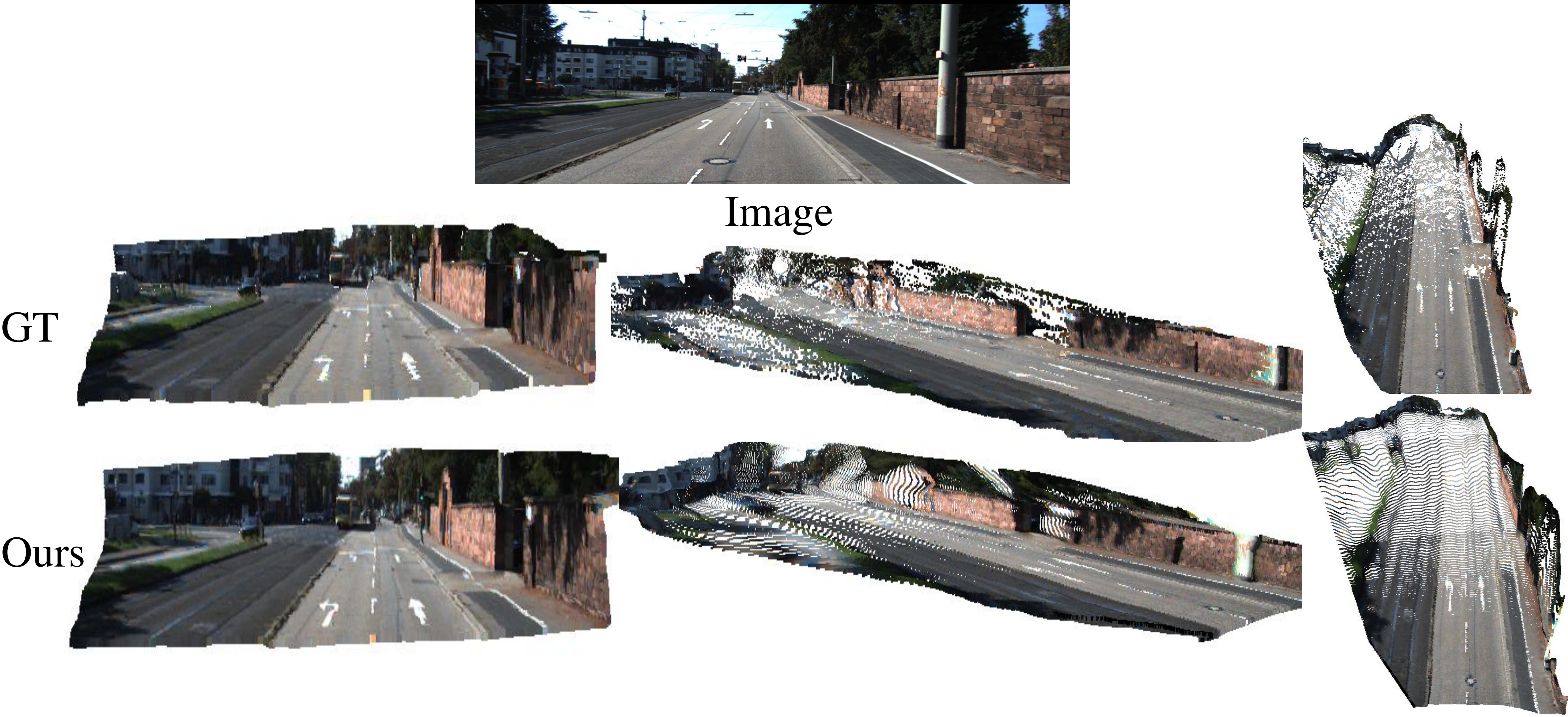}
}
\caption{Reconstructed point clouds. Three scenes are randomly selected from KITTI. For the reconstructed point cloud of each scene, 3 views are selected to demonstrate the quality of the point cloud. (a) Scene 1; (b) Scene 2; (c) Scene 3. } %
\label{fig:pcd_kitti}  %
\end{figure*}

In order to further show the quality of reconstructed point cloud from the predicted depth, we randomly select 3 scenes from the testing part of NYUD-V2 and KITTI. 3 views are randomly selected to display the reconstructed point cloud. The results are shown in Fig.~\ref{fig:pcd_nyudv2} and Fig.~\ref{fig:pcd_kitti}.

\section*{Acknowledgments}
We would like to thank Huawei Technologies
for the donation  of GPU cloud computing resources.
We are particularly grateful to
one of the reviewers who sees the value of our work and has provided
constructive comments.

{\small
\bibliographystyle{./ieee}
\bibliography{arxiv_edition}
}

\end{document}